\documentclass{article}

\usepackage{arxiv}

\usepackage[utf8]{inputenc} 
\usepackage[T1]{fontenc}    
\usepackage{hyperref}       
\usepackage{url}            
\usepackage{booktabs}       
\usepackage{amsfonts}       
\usepackage{nicefrac}       
\usepackage{microtype}      
\usepackage{lipsum}
\usepackage{graphicx}
\graphicspath{ {./images/} }

\usepackage{graphicx} 
\usepackage{multicol}
\usepackage{float}
\usepackage{arabtex}
\usepackage{utf8}
\setcode{utf8}
\usepackage[export]{adjustbox}
\usepackage{tabularx,rotating}
\usepackage{booktabs}
\usepackage{multirow}
\usepackage{caption}
\usepackage{subcaption}
\usepackage{comment}
\usepackage{rotating}
\usepackage{lscape}

\title{Content-Localization based System for Analyzing Sentiment and Hate Behaviors in Low-Resource Dialectal Arabic: English to Levantine and Gulf}

\author{
 Fatimah Alzamzami \\
  School of Electrical Engineering and Computer Science\\
  University of Ottawa\\
  Ottawa, Ontario, Canada \\
  \texttt{falza094@uottawa.ca} \\
   \And
 Abdulmotaleb El Saddik \\
  School of Electrical Engineering and Computer Science\\
  University of Ottawa\\
  Ottawa, Ontario, Canada \\
  Mohamed bin Zayed University of Artifcial Intelligence\\
  Abu Dhabi, UAE\\
  \texttt{elsaddik@uottawa.ca} \\
}

\begin{document}
\maketitle
\begin{abstract}
Even though online social movements can quickly become viral on social media, languages can be a barrier to timely monitoring and analyzing the underlying social behaviors. This is especially true for  under-resourced languages on social media like dialectal Arabic; the primary language used by Arabs on social media. Therefore, it is crucial to be mindful of this and provide cost effective solutions to efficiently exploit resources from high-resourced languages to solve language-dependent online social behavior analysis in under-resourced languages on social media. This paper proposes to localize content of resources in high-resourced languages into under-resourced Arabic dialects. Content localization goes beyond content translation that converts text from one language to another; content localization adapts culture, language nuances and regional preferences from one language to a specific target language and dialect. Automating understanding of the natural and familiar day-to-day expressions in different regions, is the key to achieve a wider analysis of online social behaviors especially for smart cities. In this paper, we utilize content-localization based neural machine translation to develop sentiment and hate speech classifiers for two low-resourced Arabic dialects: Levantine and Gulf. Not only this but we also leverage the power of unsupervised learning to facilitate the analysis of sentiment and hate speech predictions by inferring hidden insights and topics from the corresponding data and providing coherent interpretations of those insights and topics in their native language and dialects. The experimental evaluations on real data have validated the effectiveness of our proposed system in precisely distinguishing positive and negative sentiments as well as accurately identifying hate content in both Levantine and Gulf Arabic dialects. Our findings shed light on the importance of considering the unique nature of dialects within the same language and ignoring the dialectal aspect would lead to inaccurate and misleading analysis. 
In addition to our experimentation results, we present a proof-of-concept of our proposed system using COVID-19 real data collected directly from Lebanon and Saudi Arabia geo-regions. We study the sentiment and hate speech behaviors during the COVID-19 pandemic in Lebanon and Saudi Arabia and provide a comprehensive analysis from three views: temporal, topic-based, and dialect-based. Our proposed unsupervised learning methodology has shown reliability in discovering insightful topics and dynamically providing coherent phrases to interpret the inferred topics in two different Arabic dialects; Levantine and Gulf.
\footnote{\scriptsize \textbf{Disclaimer:} This work uses terms, sentences, or language that are considered foul or offensive by some readers. Owing to the topic studied in this thesis, quoting toxic language is academically justified but I do not endorse the use of these contents of the quotes. Likewise, the quotes do not represent my opinions and I condemn online toxic language.}
\end{abstract}


\section{Introduction}
\label{intro}

The trend in research is that researchers tend to build data resources for every problem in every language and even dialects \cite{john2017sentiment, hate-hind, alruily2020sentiment, alzamzami2021monitoring}. This can be seen in the huge discrepancy of resources between languages, where very few have a high-resource status while many others are low-resourced. English, for instance, is a high-resourced language while $> 70\%$ of OSN users speak languages other than English, of which dialectal Arabic is among the top used languages on social media; yet it is considered a low-resourced language. This research trend limits its current systems to generalize to other languages. Moreover, the expensive cost (i.e. in terms of HW/SW requirements, time, efforts, cost, and human labor) for building a data resource for each language and dialect has definitely contributed to the under-resourced status of languages like dialectal Arabic; the primary language used on social media among Arabs. Language under-resource issue introduces a significant barrier to smart city  authorities who need access to large data volumes containing important information about the online social behaviors (OSB) of citizens. Understanding the online social behaviors is crucial for making informed decisions about how to improve and manage smart cities. One way to address the language under-resource issue is to develop intelligent tools that allow communication between high and low resourced languages so that it would make it possible to exploit existing data resources to solve OSB tasks in low-resourced languages. 
Machine translation is a common tool to bridge the communication gap between high and low resourced languages on social media; however, word-to-word translation does not ensure transferring the context, culture, and tone of messages from a language/dialect to another language/dialect, which, in turn, might not resonate with the familiar social day-to-day expressions tailored to specific languages/dialects. In other words, traditional translation approach does not ensure that the translated content is culturally accurate and appropriate to understand the cognitive, affective, and emotional aspects of online social behaviors residing within shared content on social media.
Further, conversations on social media do not follow certain rules and are dominated by informal nature which current machine translation systems are still immature at delivering accurate translations to the informal social conversations. In response, we propose content localization system that is able to understand informal communications on social media (i.e. in high-resourced languages) and transfer their context, culture, and tone to languages and dialects with low resource status. This work proposes to localizes data resources (i.e. collected, cleaned, and annotated) of a high-resourced language into a low-resourced dialectal language. The localized data resources are then used to develop OSB models (i.e. sentiment and hate speech in this paper) in the low-resourced language/dialects with minimized time, effort, and human labor requirements. In this paper, we examine the validity of our proposed system on English as a high-resourced language and two Arabic dialects (Levanitne and Gulf) as low-resourced language/dialects.

Given the heavy information flow being generated daily on social media \cite{tracing-OSN}, discovering the knowledge insights embedded within those information in real time is of great importance \cite{dst}; this is extremely crucial to facilitate the analysis of corresponding online social behaviors of citizens within smart cities especially during critical situations like pandemics. Yet, it is nearly impossible to manually monitor huge loads of online data flow \cite{alzamzami2020light}. Thanks to the unsupervised learning nature of topic modeling algorithms that makes it possible to cluster huge volume of data into meaningful insights fast and without prior human-knowledge involved. Since classical topic modeling methods are sensitive to noise, which social media data suffers from \cite{alzamzami2021monitoring}, proper NLP preprocessing tools are required to clean the data noise and only keep the informative pieces of information. Such tools do exist in high-resourced languages like English but are insufficient or non-existent in low-resourced languages like dialectal Arabic \cite{alayba2018combined}. Fortunately, unsupervised deep learning algorithms have been proven robust against data noises. In addition to the unique nature of Transformers' architecture and through the power of transfer learning, it has been easy and possible to identify insightful topics in languages with insufficient NLP preprocessing tools. Although topic modeling algorithms can be used to capture key insights that are present in OSN data \cite{alzamzami2021monitoring}, they rely on a set of top keywords to explain the inferred insights \cite{sent-tm-covid,hate-tm-covid-arabic} which cannot provide a comprehensive understanding of those insights \cite{mei2007automatic}. Phrases are preferable over single keywords to explain topics \cite{mei2007automatic}; single keywords do not offer context to relate the keywords of topics while sentences are too specific and usually focus on a single aspect of a topic. Given the free nature of conversation sharing on OSNs, finding the optimal lengths of phrases that best describe the aspects of a topic is challenging. Current phrase-extraction methods rely on a fixed sliding window which might miss important phrases with lengths other than the predefined ones. Some topics might use longer or shorter phrase expressions than the other and this is difficult to control on OSN open platforms that accommodate unstructured data format. This work targets the mentioned limitation and proposes to extract dynamic-length phrases to coherently describe inferred topics.

This study is intended to contribute to the public management by monitoring social media activities in low-resourced languages; this can be used to improve public well-being and to prevent potential social unrest in smart cities. By monitoring the behavior of online social activities, authorities can identify trends, concerns, or potential causes of social unrest. It could also be used to identify those whose incite violence as it has been shown that this type of social ties information can be inferred from OSN data \cite{tracing-OSN, dst}. All these information help authorities to address corresponding issues and instantly take proper decisions that ensure QoL in smart cities. We summarize the contributions of this paper as follows:
\begin{itemize}
    \item Design a content-localization based system for real-time monitoring of online social behaviors in low-resourced dialectal Arabic on social media.
    \item Develop a model for real-time data exploration and dynamic interpretation using unsupervised learning approach for two low-resourced Arabic dialects: Levantine and Gulf.
    \item Develop a content-localization based BERT sentiment classifier for two low-resourced Arabic dialects; Levantine and Gulf.
    \item Develop a content-localization based BERT hate classifier for two low-resourced Arabic dialects: Levantine and Gulf.
    \item Conduct a large scale analysis of online social behavior during COVID-19 pandemic in Lebanon and Saudi Arabia (i.e. two under-resourced Arabic dialects: Levantine and Gulf).
\end{itemize}

The rest of the paper is organized as follows. Section \ref{rw} presents the related work. Our proposed system is presented in Section \ref{method}. Section \ref{exp} explains
the experimental design and evaluation metrics whereas the results and analysis are discussed
in Section \ref{res}. Finally, in Section \ref{conc} we conclude our
proposed work and discuss possible future directions.

\section{Related Work}
\label{rw}

Neural machine translation (NMT) has made a significant progress in the past decade on language pairs with abundant resources like English-French and English-Spanish. Unlike English-MSA (Modern Standard Arabic) machine translation that has demonstrated remarkable performance lately \cite{alqudsi2019hybrid, alqudsi2014arabic, zakraoui2020evaluation}, the accuracy of English-dialectal Arabic \cite{ranathunga2023neural} machine translation is significantly lower that this of English-MSA. The high variability of Arabic dialects has definitely contributed to the scarcity of large parallel data resources which in turn introduces a major challenge in developing accurate NMT systems for low-resourced dialectal languages like Arabic \cite{ranathunga2023neural}. Despite being more widely used on social media than Modern Standard Arabic (MSA), dialectal Arabic machine translation from English is still in its early stages \cite{sajjad2020arabench}.
Zbib et al \cite{zbib2012machine} proposed a statistical machine translation (SMT) system from Arabic Egyptian and Levant to English. Another English-Egyptian MT system was studied by Nagoudi et al. \cite{nagoudi2021investigating} where transfer learning was implemented during MT training.  MDC corpus for English-Levantine/North\-African/Egyptian was proposed by Bouamor et al \cite{bouamor2014multidialectal}; a preliminary analysis of the corpus confirms the differences between dialects especially between Western and Eastern Arabic dialects. In later work, Bouamor et al \cite{bouamor2018madar} proposed MADAR parallel corpus and lexicon for city-level dialectal Arabic in the travel domain. They found that the average similarity between Arabic dialects in their dataset is 25.8\%. Qatari-English speech corpus consisting of 14.7k pair sentences was collected by Elmahdy et al. \cite{elmahdy2014development} from Qatari TV shows. The Bible \footnote{\label{bible1} https://www.biblesociety.ma}, \footnote{\label{bible2} https://www.bible.com} was translated from English to Arabic North\_African dialect. Sajjad et al. \cite{sajjad2020arabench} combined the abovementioned datasets to develop an NMT system from dialectal Arabic to English. We find that these studies suffer from at least one of eight limitations that contradict with the main objective of this study: (1) domain-dependent translation corpus, (2) inconsideration of social media communication culture like abbreviations and composed words, (3) inconsideration of idiomatic expression, (5) inconsideration of code switching, (6) non-professional translators that are not bilingual native to near native in both languages, (7) inconsideration of content localization for both context and tone of messages, (8) small size corpus. In addition, most of previous works have studied and tested a translation direction from dialectal Arabic to English. Yet, literature suggests that OSB analysis yields better results when training on a native language/dialect. Our work addresses the abovementioned limitations and propose to utilize content-localization based dialectal Arabic NMT customized for social media conversations. The objective of this study is to provide a system that minimizes the expensive cost associated with the current practise of researchers where they build data resources for every OSB problem in every language and even dialect  \cite{french-tweets-dataset, baly2019arsentd, dfsmd, haddad2019t, farha2020arabic, abdul2014sana, mulki2021let}.

Deep learning approach has reliably solved the problem of insufficient NLP language-dependent preprocessing tools for under-resourced languages like dialectal Arabic. Unlike the classical topic modeling algorithms like LDA and NMF that require efforts for data preprocessing and hyperparameter tuning in order to predict meaningful clusters or topics, BERT-based (i.e. Transformer-based) topic modeling approach \cite{grootendorst2022bertopic} alleviates this requirement by leveraging pre-trained language models that learn the contextual representations of words at the presence of data noise instead of the classical way of learning on count data and ignoring the order and context of words. However, BERT-based topic models rely on the top single keywords to describe inferred topics. According to the literature \cite{mei2007automatic}, people favor the use of phrases over single keywords and sentences to describe a topic; they claim that combining keywords create difficulties to comprehend the main meaning of topics while long sentences are too specific and might miss capturing other aspects of the topic. Existing phrase extraction methods \cite{siddiqi2015keyword,mihalcea2004textrank,wan2008single, danesh2015sgrank,florescu2017positionrank} not only rely on language-dependent preprocessing tools (e.g. chuncking, POS tagging, and $n$-gram), but also rely on fixed sliding windows to extract phrases. This contradicts with the the diversity of OSN conversations (i.e. or expressions) that could be in any language/dialect of any length. In other words, a pre-defined sliding window for phrase extraction makes it difficult to determine the optimal lengths of phrases to describe topics inferred from social media data.
Further, human bias can be arise from manual interpretation of topics \cite{mei2007automatic}. Also, given the diversity and huge size of OSN contents makes the availability of domain experts to label topics a difficult task. Exploiting external knowledge resources to automatically label topics is also not applicable to the data streams of social media since emerging social contents might not exist in these external resources in a timely manner \cite{topic-labeling-twitter-summary-frameowrk}. The existing topic interpretation studies targeting social media data have been either focusing on single keywords \cite{topic-labeling-twitter-summary-frameowrk} or fixed-length phrases \cite{topic-label-likes} to describe topics resulted from topic models. The meaning of a sentence varies with the length and order of its constituting words. This paper proposes to use RAKE algorithm \cite{rake} for dynamic-length topic interpretation as it solves the mentioned issues found in  current topic labelling methods. To the best of our knowledge, this paper is the first to address these issues and to utilize RAKE algorithm for automatic interpretation of BERTopic-style topics using two under-resourced Arabic dialects on social media: Levant and Gulf.

\section{Method}
\label{method}
\begin{figure}
\centering
\includegraphics[width=.6\textwidth]{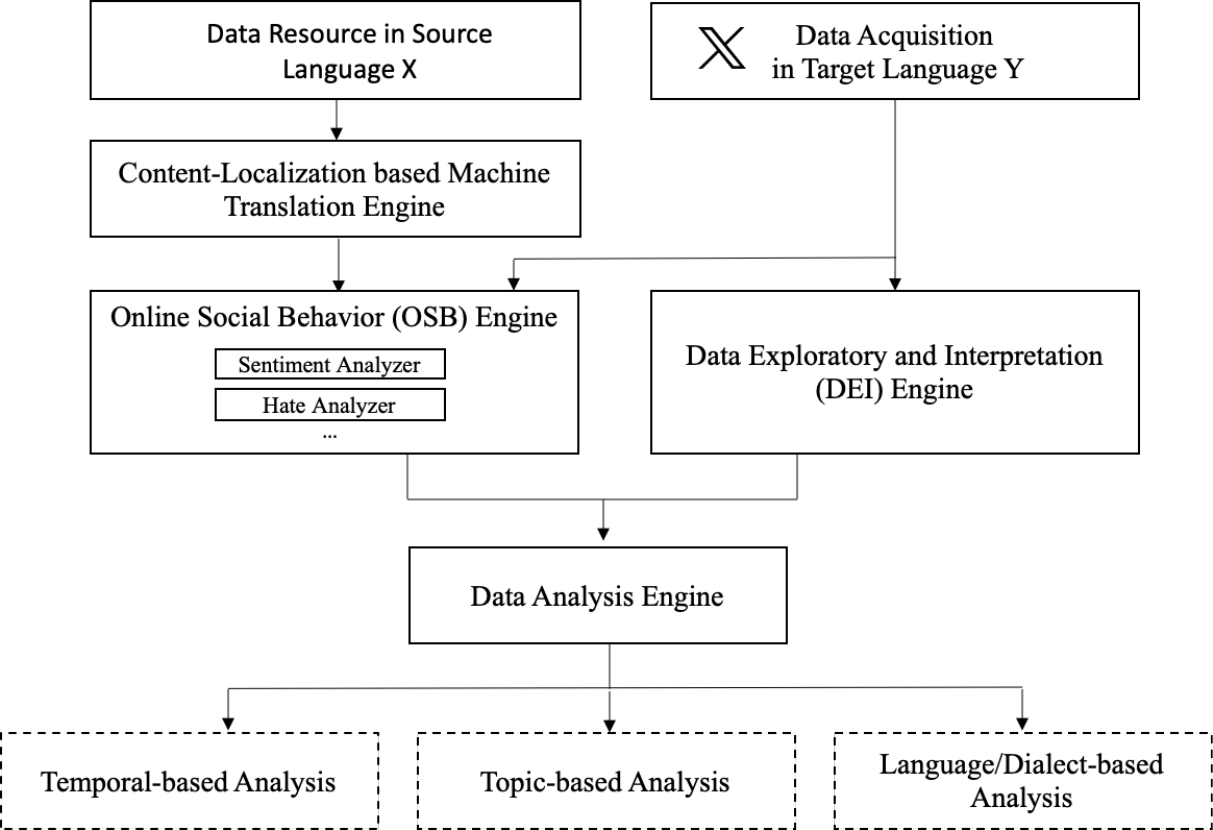}
\caption{Proposed system.}
\label{fig:framework-general}
\end{figure}

Figure. \ref{fig:framework-general} presents the proposed content-localization based system for modeling and analyzing online social behaviors (i.e. sentiment and hate speech in this work) in low-resourced Arabic dialects (i.e. Levantine and Gulf in this study). An online social behavior (OSB) data resource in a high-resourced language is localized to a low-resourced language/dialect of interest in the content-localization engine. The localized data resource is fed into the OSB engine to develop the OSB models of interest (i.e. sentiment and hate analyzers in this paper) in the target language/dialect. Supervised learning approach is used to train and test the OSB models. 
Data-to-be-analyzed (i.e. COVID-19 case study) is collected in the language/dialect of interest from social media platform (i.e. Twitter in this paper), and then fed in parallel into the OSB engine and data exploratory and interpretation (DEI) engine. The models in the OSB engine produce the corresponding predictions (i.e. sentiment and hate) from the data. In the DEI engine, unsupervised leaning algorithms are used to develop the DEI models.
The data analysis engine uses the outputs of both OSB and DEI engines
and creates an analytic story from three views: language/dialect-based, temporal, and topic-based analysis.
Language/dialect-based analysis provides a cultural view of online social behaviors. In temporal analysis, the online social behavior is illustrated in a time-line manner (i.e. over
days, months, etc) whereas topic-based analysis provides analysis based on the themes inferred from the given data.

\subsection{Content-Localization based Neural Machine Translation}
Content localization goes beyond translation which converts messages from a language to another. Content localization adapts the content and context of messages to a specific language by taking into consideration local terminology and cultural customs. Same words might convey different meaning in different dialects; for example, the word "{\scriptsize \<صاحبي>}" in Arabic-Gulf dialect refers to a friend while in Arabic-Lebanese dialect it refers to boyfriend.

 \begin{figure}[th]
\centering
\includegraphics[width=.6\textwidth]{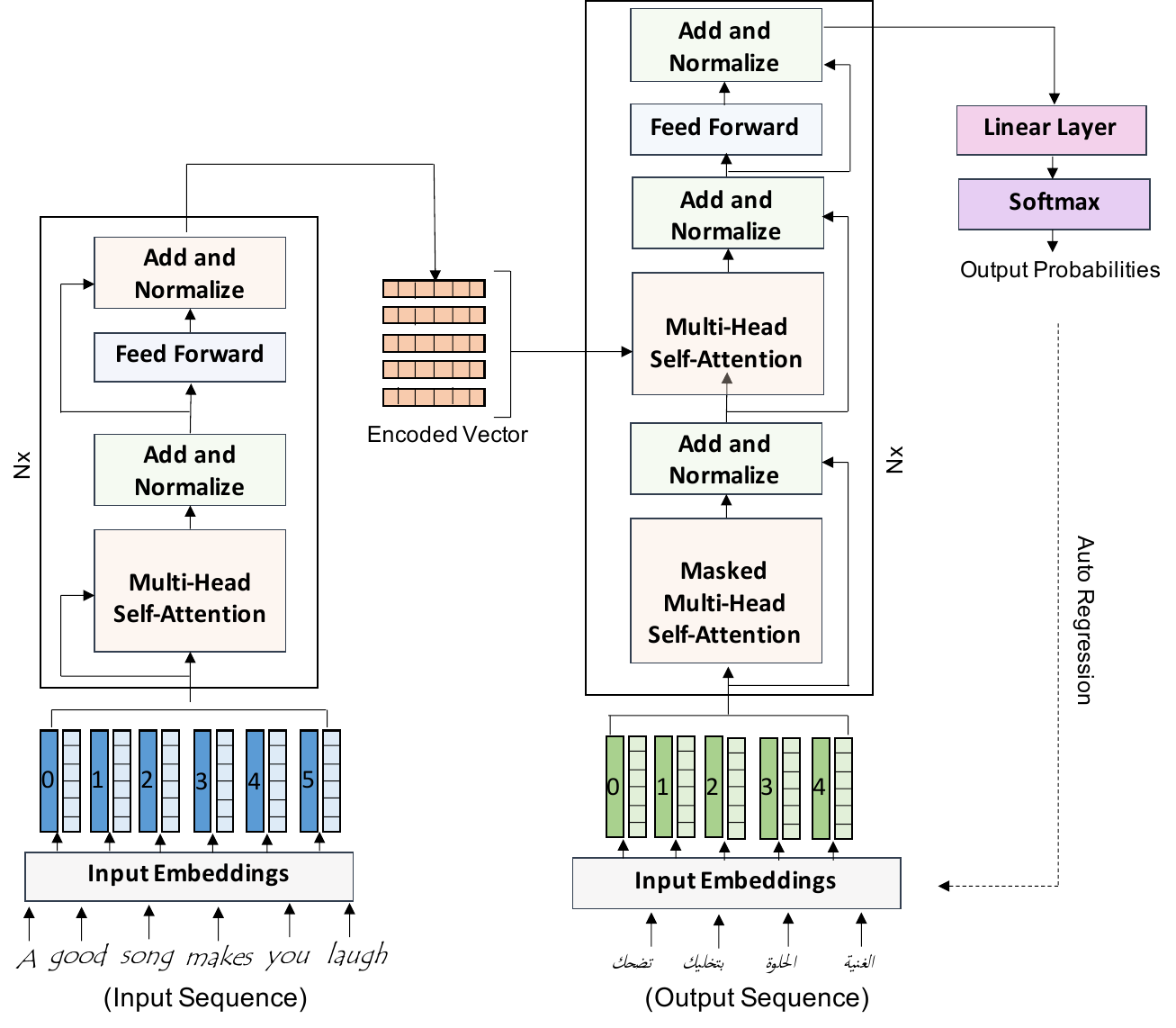}
\caption{Sequence-to-sequence Transformers architecture used for training the content-localization based NMT models adopted in this work.}
\label{fig:nmt-arch}
\end{figure}

In this work, we adopt the content localization translation approach to localize existing data resources (i.e. cleaned and annotated) in high-resourced languages (i.e. English in this paper) into low-resourced languages (i.e. dialectal Arabic in this paper) as an attempt to expand the analysis of online social behaviors across languages/dialects without the burden of creating new data resources for every language/dialect. It is important to mention that constructing new data resources is expensive in terms of time, efforts, cost, and human labor. Not only this, but recruiting and retaining domain experts or workers who are willing to commit to annotating large datasets is also a barrier. Two neural machine translation models \cite{MDAD-arx} are utilized to localize annotated sentiment and hate speech datasets in English (i.e. high-resourced language) into two low-resourced Arabic dialects; Levantine and Gulf. It is to note that Transformer architecture has been used to train the NMT models. Further, the development of the NMT models has considered five criteria \cite{MDAD-arx}: (1) Content localization based translation approach, (2) Consideration of OSN cultural language and expressions: slang abbreviations, iconic emotion (e.g. emojis, emoticons) should be kept in the
localized texts while preserving their order and context; hashtag
words (single-word and composed-word hashtags) are also localized into the corresponding Arabic dialects, (3) Consideration of informal Language: informal language is used in daily conversations; it includes slang expressions like "lol", "OMG, idk but I'm feeling down today", (4) Consideration of idiomatic expressions: an idiomatic expression should not be word-to-word translated; instead, it should be localized to convey the context or its equivalent idiomatic expression in the corresponding dialects, (5) Consideration of language code borrowing: code borrowing refers to the use of one primary language but mixing in words from another language to fit the primary language. For instance, the word "lol" is written using the Arabic alphabets as
"{\scriptsize \<لول>}"
 ; similarly, the word "cheese" is written using the Arabic alphabets as
"{\scriptsize \<تشيز>}".  Figure. \ref{fig:nmt-arch} illustrates the Transformers architecture used for the development of the content-localization based NMT models; it consists of 12 layers of encoder and 12 layers of decoders with model dimension of 1024 on 16 heads. On top of both encoder and decoder, there is an additional normalization layer that was found to stabilize the training. Weights were initialized from mBART pre-trained model \cite{liu2020multilingual} that was trained on 25 distinct languages. The NMT models were finetuned on our English-multidialectal Arabic dataset customized for social media conversations \cite{MDAD-arx}.

\subsection{Online Social Behavior (OSB) Modeling}

This paper studies two types of online social behaviors: sentiment and hate speech. The general supervised deep learning classification framework is followed to build our sentiment and hate classifiers. Data is prepared and preprocessed first before the training process starts. BERT architecture is utilized in training our OSB classifiers. A BERT pre-trained model is fine-tuned by training the entire BERT architecture on our localized datasets in order to alleviate any possible biases resulted from the pre-training \cite{rietzler2019adapt,alzamzami2023transformer}. BERT-base-arabic-camelbert-mix model \cite{inoue-etal-2021-interplay}, used in this work, is 
 pre-trained on a mix of Modern Standard Arabic (MSA), dialectal Arabic (DA), and classical Arabic (CA).
A classification layer is appended to the BERT layer where logits are produced. Soft max layer is used to normalize the output logits and computes the probability of classes. The optimizer used is Adam with a learning rate of 1e-4, a weight decay of 0.01, learning rate warmup for 10,000 steps and linear decay of the learning rate after. Early-stopping approach is used to avoid over-fitting the neural network on the training data and improve the generalization of the models. The models are optimized using Adams optimizer. Finally, the models are evaluated and tested using validation and test sets before producing the final predictions. The prediction of sentiment analyzer is one of two classes: positive or negative sentiments, whereas the hate analyzer predicts one of two classes: hate or non-hate.

\subsection{Data Exploration and Interpretation (DEI)}

 \begin{figure}[h]
\centering
\includegraphics[width=.7\textwidth]{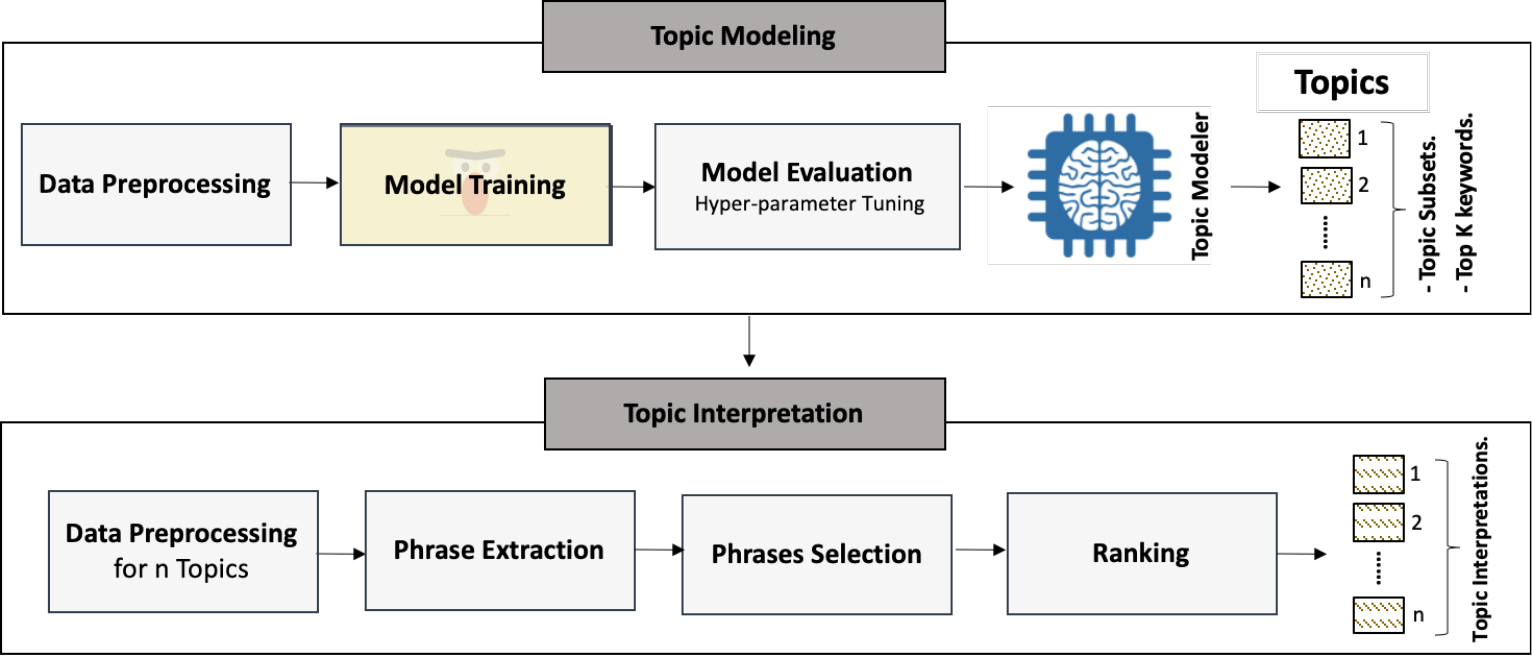}
\caption{Proposed methodology for data exploration and interpretation.}
\label{fig:framework-tm}
\end{figure}

The DEI component is mainly responsible for exploring social media data by intelligently finding  major themes in the data and then generating explainable interpenetration of these themes. Topic modeling is the approach adopted in this paper to explore and discover latent patterns (i.e. themes or topic) within large datasets. The general framework used in unsupervised topic learning is followed in this paper; Figure. \ref{fig:framework-tm} illustrates the methodology used to build our topic model. 
The data preprocessing of topic modeling is implemented based on the criteria to increase the topic relevance and minimize the noise (i.e uninformative) parts of the data. To learn the topics, BERT-based (i.e. Transformer-based) \cite{grootendorst2022bertopic} is used in this study. BERT-based topic modeling leverages pre-trained large language models that learn the contextual representations of word features unlike traditional topic modeling approach (e.g. LDA and NMF) that learn on count data ignoring the context and order of words \cite{alzamzami2021monitoring}. The output of the topic model will be $k$ topics. For each topic, we consider the top $n$ keywords and their corresponding subsets of data documents (i.e. tweets in this study). 

After the topic clusters have been inferred, they are fed into the topic interpretation component to automatically generate the interpretations of the topics; hence giving us a deeper understanding of the topics without manual intervention. Using the top $n$ keywords only is inadequate in interpreting the coherent meaning of topics \cite{mei2007automatic}.
A good interpretation of a topic should be able to capture its meaning and distinguish it from other topics. Single words are not able do this as they lack the context of phrases and sentences \cite{mei2007automatic}.
To elaborate, single keywords are too broad and may overlook the semantic relationships needed to convey the main idea of the topics. Phrases, on the other hand, provide context to single words, resulting in stronger topic cohesion. Additionally, phrases are inherently broad, allowing them to capture the overall meaning of topics \cite{mei2007automatic}. 
This paper proposes the use of an unsupervised dynamic-length phrase extraction approach; RAKE -Automatic Rapid Keywords Extraction- algorithm \cite{rake} which is utilized to identify topic phrases. 
Note that RAKE algorithm does not require specific domain or language dependencies; it is domain and language independent keyword-extraction algorithm that uses word frequency and co-occurrence to identify meaningful phrases.
RAKE is a graph-based algorithm that was designed based on the assumption that the multiple words that constitute a keyword are rarely split by stop words or punctuation marks. The assumption states that stop words and punctuation are uninformative unlike the remaining words that are assumed to be informative; they are referred to as content words. RAKE starts extracting phrases by splitting a given text into a set of candidate keywords at the occurrence of word delimiters. Next, the resulted set of candidate keywords is split into a sequence of consecutive words at the occurrence of phrase delimiters or stop words. The consecutive words within a sequence together form a new candidate keyword (i.e. phrase). A word-word co-occurrence graph is created to be later used in computing the scores of the candidate phrases. Three scoring metrics were proposed: (1) $deg(word)$; word degree to calculate  words with frequent occurrences in a given document as well as in longer candidate phrases, (2) $freq(word)$: word frequency which computes frequent words without taking into consideration the word-word co-occurences, (3) $\frac{deg(word)}{freq(word)}$: ratio of degree to frequency. In this work, we use the degree of word metric $deg(word)$ to calculate phrases scores. The score of a candidate phrase is computed as the sum of its constituting words' scores.
RAKE algorithm has been proven computationally fast and effective in dynamic-length phrase extraction from limited-sized social media datasets for automatic and coherent topic interpretations \cite{alzamzami2021monitoring}.

We follow the methodology depicted in Figure. \ref{fig:framework-tm} to extract phrases of topics. First, data subsets of the inferred topics are preprocessed; each is preprocessed independently. This process is similar to that of topic modeling; however, in phrases extraction stopwords are kept. Second, RAKE algorithm is applied to each topic data subset independently to extract corresponding keywords and phrases. The weights of the resulted keywords and phrases are calculated using the degree of word metric $deg(word)$ that computes the words that have frequent occurrences in a document as well as in longer candidate phrases.  Third, keywords and phrases based on the top $n$ keywords (i.e. resulting from our BERT-based topic model) are selected. Before selecting RAKE keywords and phrases, the duplicated keywords (i.e. resulting from our BERT-based topic model) across topics are removed. Keyword duplication might lead to ambiguous interpretations of topics. By removing duplication, we ensure that each topic is interpreted in a distinctive way. Finally, the keywords and phrases are assigned weights based on their importance (i.e. degrees), and then ranked according to those degrees. The output phrases have different lengths, with a minimum of two words. To determine the optimum lengths of phrases, the average lengths of phrases for each topic is calculated. After calculating the average, phrases with more general concept and phrases with more specific concept than the average are considered. This is done by selecting shorter phrases and longer phrases than the calculated average length.

\section{Experiment Design and Evaluation Metrics}
\label{exp}

\subsection{Datasets and Preprocessing}

We list below the datasets we have used for modeling and evaluating our proposed system: \\
\textbf{SemEval-2013/2017 Sentiment Dataset \cite{nakov-etal-2013-semeval, rosenthal2017semeval}}: We have combined the two sentiment datasets of SemEval 2013 and 2017, and removed the neutral class. We have balanced the classes to obtain 6,500 positive tweets and 4,523 negative tweets. This dataset is localized into Arabic Levantine and Gulf using our proposed NMT models to be used later for sentiment modeling.\\
\textbf{ArSentD-Lev \cite{baly2019arsentd}}:
This dataset consists of 1,232 positive tweets and 1,884 negative tweets collected from the Arabic Levant region, and manually annotated through the crowd-sourcing approach. This dataset is used for evaluation purposes.\\
\textbf{OCLAR datasets \cite{omari2022oclar}}:
OCLAR is an opinion corpus for Arabic Lebanese reviews. The positive class is the reviews rating from 1 to 3 (3,465 reviews), while the negative class is the reviews rating from 1 to 2 (451 reviews). This dataset is used for evaluation purposes.\\
\textbf{Saudi Banks Dataset \cite{alqahtani2022customer}}: 
This manually annotated dataset contains Arabic-Saudi tweets from four Saudi banks. The dataset contains 8,669 negative tweets and 2,143 positive tweets. This dataset is used for evaluation purposes.\\
\textbf{Saudi Vision-2030 Dataset \cite{alyami2020application}}: 
This  manually annotated dataset contains tweets discussing several aspects of Saudi Vision 2030. It consists of 2,436 positive tweets and 1,816 negative tweets. This dataset is used for evaluation purposes.\\
\textbf{HatEval Dataset \cite{hateVal-ds}}: This manually annotated and approved English dataset has been constructed based on women or immigrants as targets of hate speech. The dataset contains 5,470 hate tweets. This dataset is translated into Arabic Levantine and Gulf using our proposed NMT models to be used later for hate speech modeling.\\
\textbf{Let-Mi datasets \cite{mulki2021let}}: 
This dataset consists of Levantine tweets annotated for detecting misogynistic behavior on online social media. This dataset, which consists of 2,654 hate tweets and 2,586 non-hate tweets, has been annotated manually by Levant people. This dataset is used for evaluation purposes.\\
\textbf{COVID-19 datasets - Lebanon \cite{zahra2020targeted}}:
Arabic tweets were retrieved using geo-coordinates of Lebanon during the COVID-19 pandemic in 2020. This dataset is used for evaluation purposes in our COVID-19 case study.\\
\textbf{COVID-19 datasets - Saudi Arabia \cite{saudi-covid19-dataset}}: 
Arabic COVID-19 related tweets were collected using geo-coordinates of Saudi Arabia during eight consecutive days; 14-21 of March 2020. This dataset is used for evaluation purposes in our COVID-19 case study.

A list of preprocessign steps have been implemented to prepare the data before the modeling the stage: (1) Removing extra whitespaces, (2) Removing encoding symbols, (3) Removing URLs, (4) Converting text to lower case, (5) Removing tashkeel and harakat: tashkeel or harakat refer to all the diacritics placed over or below letters, (6) Normalizing Hamza, (7) Removing user mention, (8) Removing special characters and numbers, (9) Removing stopwords.

\subsection{Experimental Design and Evaluation Protocol}

We design the experiments of this study as follows:\\
(1) We localize existing annotated sentiment datasets in English into a target language/dialect (i.e Arabic Levantine and Gulf in this experiment). Note that we preserve the source annotations of the source datasets. We then train sentiment classifiers using these localized datasets. Later, we evaluate our trained classifiers on external datasets under the condition that the external datasets should have been created and annotated in the native target language and dialects. In this experiment, we localize the English sentiment datasets \cite{nakov-etal-2013-semeval, rosenthal2017semeval} to Arabic Levantine and Gulf using our proposed NMT models. The localized datasets will be used later to train an Arabic sentiment classifiers for Levantine and Gulf dialects. The purpose of this experiment is to examine the validity of our proposed approach that aims to minimize the dependecy of language/dialects in modeling online social behavior (i.e. sentiment in this experiment), especially in low-resourced languages/dialects.\\
(2) We localize an existing annotated English hate speech dataset into two target Arabic dialects, Levantine and Gulf, using our proposed NMT models. Then, we train two Arabic hate classifiers using these localized datasets, one for each dialect. Note that we preserve the source annotations of the source dataset.  After evaluating our dialectal Arabic hate classifiers on validation split, we further assess their performance on an external dataset of Arabic Levantine dialect. The Levant external dataset has been constructed and annotated in the native Levant dialect by native Levant speakers. In this experiment, we localize the English hate dataset \cite{hateVal-ds} to Arabic Levantine and Gulf dialects to be used later for training Arabic-Levantine and Arabic-Gulf hate classifiers. The purpose of this experiment is to examine the impact of different dialects of the very same language on the analysis of online social behaviors on social media (i.e. hate speech in this experiment). \\
For sentiment and hate speech modeling, the datasets have been randomly split into 80\% for training and 20\% for validation. We evaluate our models on the validation splits during training- every 100 steps- to track their learning progress. We have used early-stopping approach to prevent any potential overfitting to the training data by regularizing the model learning during the training process. Accuracy, precision, recall, and F-score, are common evaluation metrics used for supervised classification evaluation. It is worth noting that precision, recall and f-score give a better view of model performance than accuracy alone does.\\
(3) We use Transformer-based deep topic learning to train and evaluate our topic models using two dialectal Arabic datasets; COVID-19 datasets for Lebanon \cite{zahra2020targeted} (i.e. Levant dialect) and Saudi Arabia \cite{saudi-covid19-dataset} (Gulf dialect).
BERTtopic \cite{grootendorst2022bertopic} is used to learn representative topics from the two COVID-19 datasets. Pre-trained Arabic language model is used as embeddings. Note that it is not needed to define the number of topics in advance as BERTopic \cite{grootendorst2022bertopic} uses HDBSCAN clustering algorithm that does not allow to pre-define the number of clusters. Coherence score \cite{coherence-score} is used as a metric to evaluate the performance of our topic models. A coherence score for a topic is calculated by measuring the degree of semantic similarity between high scored words within the topic.\\
(4) For topic dynamic interpretation (i.e. phrase extraction), we implement RAKE algorithm and apply it to each tweet subset corresponding to each topic inferred by our BERTopic models. We found out that the average phrase length of three has the highest frequency in all the topics for Lebanon and Saudi Arabia datasets. Therefore, we decide on considering phrases of lengths two and four to add a more general meaning and a more specific meaning to the phrases of length three.

\section{Results and Analysis}
\label{res}

\subsection{Topic Modeling and Interpretation}
\begin{figure*}[t!]
\centering
\includegraphics[width=0.98\textwidth, height=40mm]{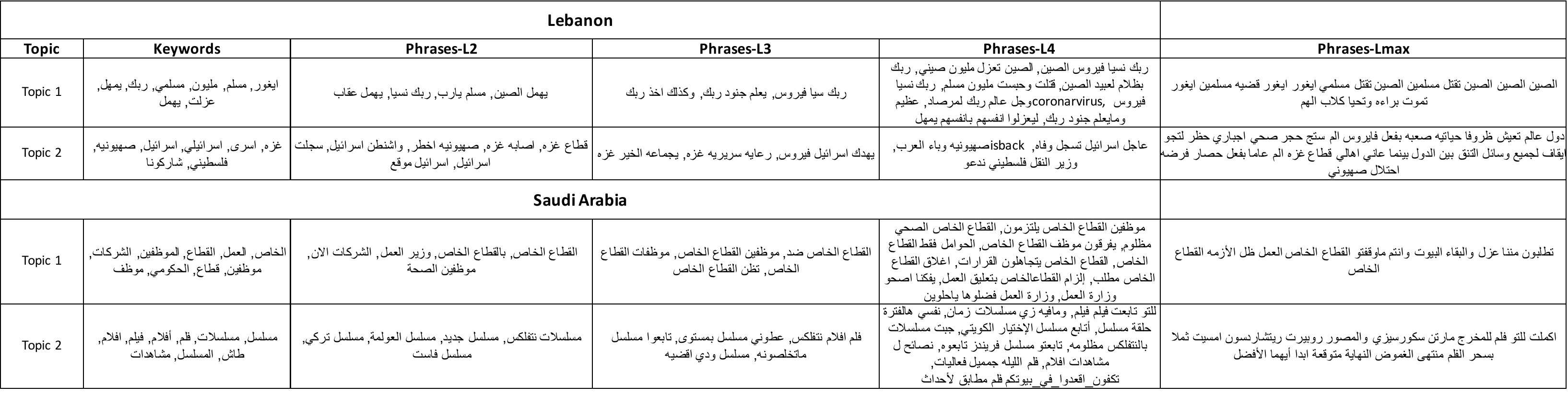}
\caption{Top keywords and phrases extracted using RAKE based on BERTopic top keywords. The keywords and phrases are ranked based on RAKE - Lebanon and Saudi Arabia.}
\label{fig:rake-phrases-leb-saudi}
\end{figure*}

Two BERTopic-based models that have been trained on two Arabic datasets; COVID-19 in Lebanon (i.e. Arabic-Levantine dialect) and COVID-19 in Saudi Arabia (i.e. Arabic-Gulf dialect). Both has performed at 0.1 of coherence score on topic size of 100 that has been determined by BERTopic during the training process. We group the inferred topics for both Lebanon and Saudi Arabia into categories as seen in Figure. \ref{fig:leb-saudi-sent-overall-topics}. Figure. \ref{fig:rake-phrases-leb-saudi} lists a sample of the top 5 keywords inferred by our BERTopic models; the top five keywords are shown to provide a general idea about the inferred topics but not a coherent interpretations of topics.

In Figure. \ref{fig:rake-phrases-leb-saudi}, we list the results of our RAKE-based phrases extracted from COVID-19 datasets of Lebanon and Saudi Arabia. The figure shows a sample of topic keywords and phrases used during COVID-19 pandemic over there. 
Note that  a larger topic samples with their keywords and phrase interpretations are presented in the appendix in Figure.\ref{fig:rake-phrases-leb-saudi-appendix}. By looking at the keywords of topic 1-Saudi Arabia (Figure.  \ref{fig:rake-phrases-leb-saudi}), we can see  that these keywords are  mainly about the private sector: "{\scriptsize \<القطاع الخاص>}" meaning private sector,
 "{\scriptsize \<العمل>}"  work,
 "{\scriptsize \<القطاع>}"  sector,
 "{\scriptsize \<الموظغين>}"  employees,
 "{\scriptsize \<الشركات>}"  companies,
 "{\scriptsize \<الحكومي>}"  governmental, and 
 "{\scriptsize \<موظف>}" an employee. 
However, this set of single keywords is not contextualized, therefore, the message is not conveyed.  Thanks to our topic interpreter that provides details about the keywords by adding contexts which have facilitated the understanding of the topic. Phrases of length 3 and 4 have added some information about the private sector and employees; the phrases of length 3 
 ("{\scriptsize \<القطاع الخاص ضد>}" meaning private sector is against,
"{\scriptsize \<موظفين القطاع الخاص>}" private sector's male employees,
"{\scriptsize \<موظفات القطاع الخاص>}" private sector's female employees)
indicate that the private sector is against its employees of both genders male and female. The idea of the topic is wrapped up in phrases of length 4; the phrases  
"{\scriptsize \< القطاع الخاص الصحي مظلوم>}" meaning the private health sector is oppressed, 
 "{\scriptsize \<القطاع الخاص يتجاهلون القرارات>}" private sector is ignoring the decisions, 
 "{\scriptsize \<اغلاق القطاع الخاص مطلب>}" shutting down the private sector is a requirement,  
 "{\scriptsize \<إلزام القطاع{\_}الخاص بتعليق العمل>}" forcing the private sector to suspend work, have added more context to complete the idea of the topic; the employees of private sectors seem to have been complaining about the private sector not abiding by the corono virus measures implemented by the authorities during the early stages of the pandemic back in 2020, and demanding that the employees go to the work place instead of quarantining at home. Similar to topic 1-Saudi Arabia, the phrases of length 2, 3 and 4 ("{\scriptsize \<مسلسلات نتفلكس>}" meaning Netflix series,
 "{\scriptsize \<مسلسل جديد>}"  a new tv series, 
 "{\scriptsize \<مسلسل فاست>}"  the Fast tv series, 
 "{\scriptsize \<مسلسل تركي>}"  a Turkish series, 
 "{\scriptsize \<افلام نتفلكس>}"  Netflix movies,  
 "{\scriptsize \<مافيه زي مسلسلات زمان>}"  nothing like old tv series, 
 "{\scriptsize \<جبت مسلسلات بالنتفلكس مظلومه>}"  I found Netflix series that are underrated, 
 "{\scriptsize \<نصائح ل مشاهدات افلام>}"  recommendations of movies to watch, 
 "{\scriptsize \<فلم الليله جمميل فعاليات>}"  the movie of tonight nice activities), have wrapped up the context of topic 2-Saudi Arabia, whose single keywords set includes  "{\scriptsize \<مسلسلات>}"  meaning tv series,
 "{\scriptsize \<فلم>}"  a movie,
 "{\scriptsize \<طاش>}"  Tash is an 80's Saudi comedy show, 
 and "{\scriptsize \<مشاهدات>}"  views. Unlike the single keywords that have failed to convey the message, those phrases have manifested that some people in Saudi have been watching globalized movies and series, giving recommendations, and voicing out their opinions on movies.

 Phrases of maximum lengths are mostly discarded since long phrases (i.e. sentences) do not provide the overall meaning of the topics. instead they only show one part of the whole topic. A case example can be seen in topic 2 of Saudi Arabia (Table. \ref{fig:rake-phrases-leb-saudi}). The long sentence 
"{\scriptsize \<اكملت للتو فلم للمخرج مارتن سكورسيزي والمصور روبيرت ريتشاردسون امسيت ثملا بسحر >\\
\< الفلم منتهى الغموض النهاية غير متوقعة ابدا أيهما الأفضل>}"  meaning 
 "I just finished watching a movie directed  by Martin Scorsese and filmed by Robert Richardson. I was enchanted by the thrilling and unpredictable aspect of the movie...", provides a specific sub-detail about the topic which shows only one part of the whole topic. 
 This finding is supported by the results obtained by Qiaozhu \cite{mei2007automatic} that sentences might not be accurate to capture the general meaning of a topic as they might be too specific. More examples of our extracted phrases can be seen in the Appendix section (Figure. \ref{fig:rake-phrases-leb-saudi-appendix}).

\subsection{Sentiment Analysis}
The results in Table. \ref{tab:mlmd-hate-sent-val} presents the performance of the proposed sentiment and hate classifiers trained using the localized dataset (i.e. translated from English to Arabic Levantine and Gulf dialects using the proposed NMT models). The results represent the classification performance using the validation split of the same data that the classifiers have been trained on. 
The English to Arabic-Levantine and English to Arabic-Gulf are shown to have effectively learnt to distinguish between positive and negative sentiment classes using the localized datasets (i.e. by the proposed NMT models).
Both Levant and Gulf sentiment classifiers have performed the same in terms of accuracy (86\%), precision (0.86), and recall (0.86)

\begin{figure}[h]
\centering
\begin{subfigure}[b]{.3\textwidth}
  \centering
  \includegraphics[width=\linewidth]{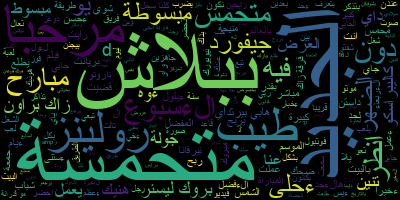}
  \caption{Positive class, Levantine model.}
  \label{fig:wc-sent-pos-model1}
\end{subfigure} 
\begin{subfigure}[b]{.3\textwidth}
  \centering
  \includegraphics[width=\linewidth]{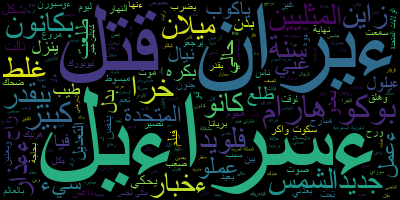}
  \caption{Negative class, Levantine model.}
  \label{fig:wc-sent-neg-model1}
\end{subfigure}\\
\begin{subfigure}[b]{.3\textwidth}
  \centering
  \includegraphics[width=\linewidth]{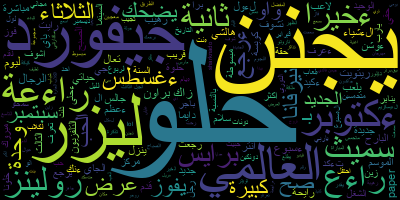}
  \caption{Positive class, Gulf model.}
  \label{fig:wc-sent-pos-model2}
\end{subfigure} 
\begin{subfigure}[b]{.3\textwidth}
  \centering
  \includegraphics[width=\linewidth]{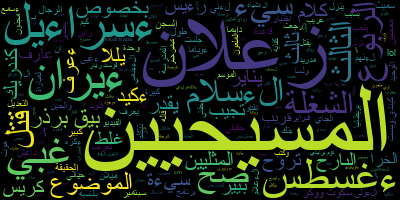}
  \caption{Negative class, Gulf model.}
  \label{fig:wc-sent-neg-model2}
\end{subfigure} 
\caption{Word-cloud generated from messages classified as positive or negative sentiment by our localized Arabic-Levantine and Arabic-Gulf sentiment models.}
\label{fig:wc-sent-en}
\end{figure}

Figure. \ref{fig:wc-sent-en} illustrates the high frequency words predicted as positive or negative classes by our sentiment classifiers. The words in the figures corresponding to the positive class (\ref{fig:wc-sent-pos-model1}, \ref{fig:wc-sent-pos-model2}) reflect positive sentiment such as "{\scriptsize \<متحمس>}" meaning "excited", "{\scriptsize \<طيب>}" meaning "good", "{\scriptsize \<يحنن>}" meaning "so good or spectacular", "{\scriptsize \<يضحك>}" meaning "funny", "{\scriptsize \<مبسوط>}" meaning "happy", "{\scriptsize \<المفضل>}" meaning "favourite", "{\scriptsize \<رائعة>}" meaning "spectacular", "{\scriptsize \<زين>}" meaning "nice". Figures related to the negative class (\ref{fig:wc-sent-neg-model1}, \ref{fig:wc-sent-neg-model2}) also reflect negative sentiment such as "{\scriptsize \<قتل>}" meaning "killing or murder", "{\scriptsize \<غلط>}" meaning "wrong or mistake", "{\scriptsize \<سيء>}" meaning "bad", "{\scriptsize \<يضرب>}" meaning "beat", "{\scriptsize \<أكره>}" meaning "I hate", "{\scriptsize \<زعلان>}" meaning "sad or upset", "{\scriptsize \<موت>}" meaning "death", "{\scriptsize \<مات>}" meaning "died", "{\scriptsize \<غبي>}" meaning "stupid or dumb". From this, we claim that our sentiment classifiers have been able to effectively learn sentiment from our localized data and to successfully distinguish between positive and negative classes which, in turn, proves the effectiveness of using the content-localization based NMT approach to transfer the context of social media texts from high-resourced language to low-resourced language and dialects. 

\begin{table*}
\caption{The performance of our localized sentiment and hate classifiers on the validation set in terms of accuracy, precision, and recall. The classifiers represent the localized Arabic-Levantine and Arabic-Gulf dialects.}
\centering
\resizebox{.7\textwidth}{!}{%
\begin{tabular}{c|c|p{7.165em}|c|c|c|c}
\toprule
& \multicolumn{1}{c}{} & \multicolumn{1}{c|}{} & \multicolumn{1}{p{4.835em}|}{\textbf{Validation \newline{}Precision}} & \multicolumn{1}{p{4.665em}|}{\textbf{Validation Recall}} & \multicolumn{1}{p{4.165em}|}{\textbf{Validation F-Score}} & \multicolumn{1}{p{4.415em}}{\textbf{Accuracy}} \\
\cmidrule{2-7}          & \multicolumn{1}{c|}{\multirow{2}[4]{*}{\textbf{Sentiment}}} & En$->$Ar-Levantine & 0.86  & 0.86  & 0.86  & 86 \\
\cmidrule{3-7}          &       & En$->$Ar-Gulf & 0.86  & 0.86  & 86    & 86 \\
\cmidrule{2-7}          & \multicolumn{1}{c|}{\multirow{2}[4]{*}{\textbf{Hate}}} & Es$->$Ar-Levantine & 0.69  & 0.68  & 0.68  & 69 \\
\cmidrule{3-7}          &       & Es$->$Ar-Gulf & 0.69  & 0.7   & 0.7   & 71 \\
    \bottomrule
    \end{tabular}
    }
  \label{tab:mlmd-hate-sent-val}
\end{table*}

Table. \ref{tab:mlmd-osb-sent-external-ds} summarizes the performance of our localized Levant and Gulf sentiment classifiers using external datasets. Each dataset that corresponds to a specific dialect is used to evaluate the classifier that has been trained on the same dialect (i.e. Levantine dataset is used to evaluate the Levantine sentiment classifier and the Gulf dataset to evaluate the Gulf sentiment classifier). The localized Arabic-Gulf sentiment classifier is evaluated on two Saudi (i.e. Gulf dialect) sentiment datasets: Saudi Bank Reviews dataset \cite{alqahtani2022customer} and Saudi Vision 2030 dataset \cite{alyami2020application}. The results show that our Gulf classifier is able to distinguish between both classes (i.e. positive and negative) in both datasets; it has performed at a positive f-score between 0.6-0.76, negative f-score between 0.7-0.93, and over all accuracy between 66\%-89\%. It can be observed that the Gulf classifier performs better on the negative class than it does on the positive class. After investigating the data, it has been found that there is mislabeling or ambiguity in some sentences as to whether they belong to the positive or negative class, as shown in the following examples:\\
- {\scriptsize "\<للامانه انا مع الاهلي ليا عشر سنين وجيتهم هارب من سامبا واستخدمت جميع المنتجات  من قروض شخصيه وعقاريه وبطايق فيزا>\\
    \< واشوفهم افضل بنك ممكن تكون تجربتك الشخصيه سيءه ولكن لا تستعجل وتروح وتورط ممكن عشان الدمج فيه شويه لخبطه>".}\\
- {\scriptsize "\<يستاهلون جميعا ساهموا في خدمه دينهم ووطنهم وليس من الانصاف اجحاف مجهودات  رجال الامن لماذا لا تقدم لهم العروض كباقي>\\
    \< منسوبي الصحه والتعليم والطيران>"}.\\
- {\scriptsize "\<يارب الوظيفة>"}.

The mislabeling and ambiguity of labeling is a common issue found in existing datasets. This in fact affects the learning and evaluation process of online social behavior modeling. It is an ongoing challenge that labeling datasets, especially for online social behavior like sentiment, still puzzles the researchers in this area \cite{li2019survey}. Overall, our localized Arabic Gulf classifier has shown a reliable performance in detecting the online sentiment behavior. Below are three examples of correct predicted sentiments:\\
- {\scriptsize "\<الامير محمد بن سلمان يحقق رؤية المملكة 2030 بدعم الشباب>".}\\
- {\scriptsize "\<هذا الرجل وضع للحق ميزان و بعدله استوى الأمير و الفقير أقنع بحربه على الفساد   وب رؤية 2030 فيها المستقبل للوطن>\\
    \< والعالم العربي والإسلامي و كافة الدول العضمى له منا الدعاء بأن يطيل الله في عمره على طاعته ليحقق ما نتطلع له من مستقبل باهر> ".}\\
- {\scriptsize "\<رؤية 2030 سوف تجعل السعوديه في مصاف الدول العضمى نثق برؤية سيدي>".}

\begin{table}[h]
  \centering
  \caption{\small The performance of our localized Arabic-Levantine and Arabic-Gulf sentiment classifiers, on external sentiment datasets, in terms of accuracy, precision, and recall.}
  \resizebox{.7\textwidth}{!}{%
    \begin{tabular}{c|p{13.415em}|c|c|c}
    \toprule
    \multicolumn{1}{r}{} & \multicolumn{1}{r|}{} & \textbf{F-Score (Positive)} & \textbf{F-Score (Negative)} & \textbf{Accuracy} \\
    \midrule
    \multirow{4}[8]{*}{\begin{sideways}\textbf{Model}\end{sideways}} &  En $->$ Ar-Gulf\newline{} evaluated on \newline{}Saudi Banks Dataset \cite{alqahtani2022customer} & 0.76  & 0.93  & 89 \\
\cmidrule{2-5}          &  En $->$ Ar-Gulf\newline{} evaluated on \newline{}Saudi Vision-2030 Dataset \cite{alyami2020application} & 0.6   & 0.7   & 66 \\
\cmidrule{2-5}          &  En $->$ Ar-Levantine\newline{} evaluated on \newline{}ArSentD-Lev \cite{baly2019arsentd} + OCLAR datasets \cite{omari2022oclar} & 0.83  & 0.82  & 83 \\
    \bottomrule
    \end{tabular}%
    }
  \label{tab:mlmd-osb-sent-external-ds}%
\end{table}%

A similar learning performance has been achieved in accurately detecting positive and negative classes by our localized Arabic-Levantine classifier on two Levantine sentiment datasets: ArSentD-Lev \cite{baly2019arsentd} and OCLAR \cite{omari2022oclar}. The localized Arabic-Levantine classifier has scored a high learning accuracy of 88\% with capability of separating positive and negative classes at positive and negative f-score of 0.83 and 0.82, respectively.

\begin{figure}[h]
\centering
\begin{subfigure}{.47\textwidth} 
  \includegraphics[width=\linewidth]{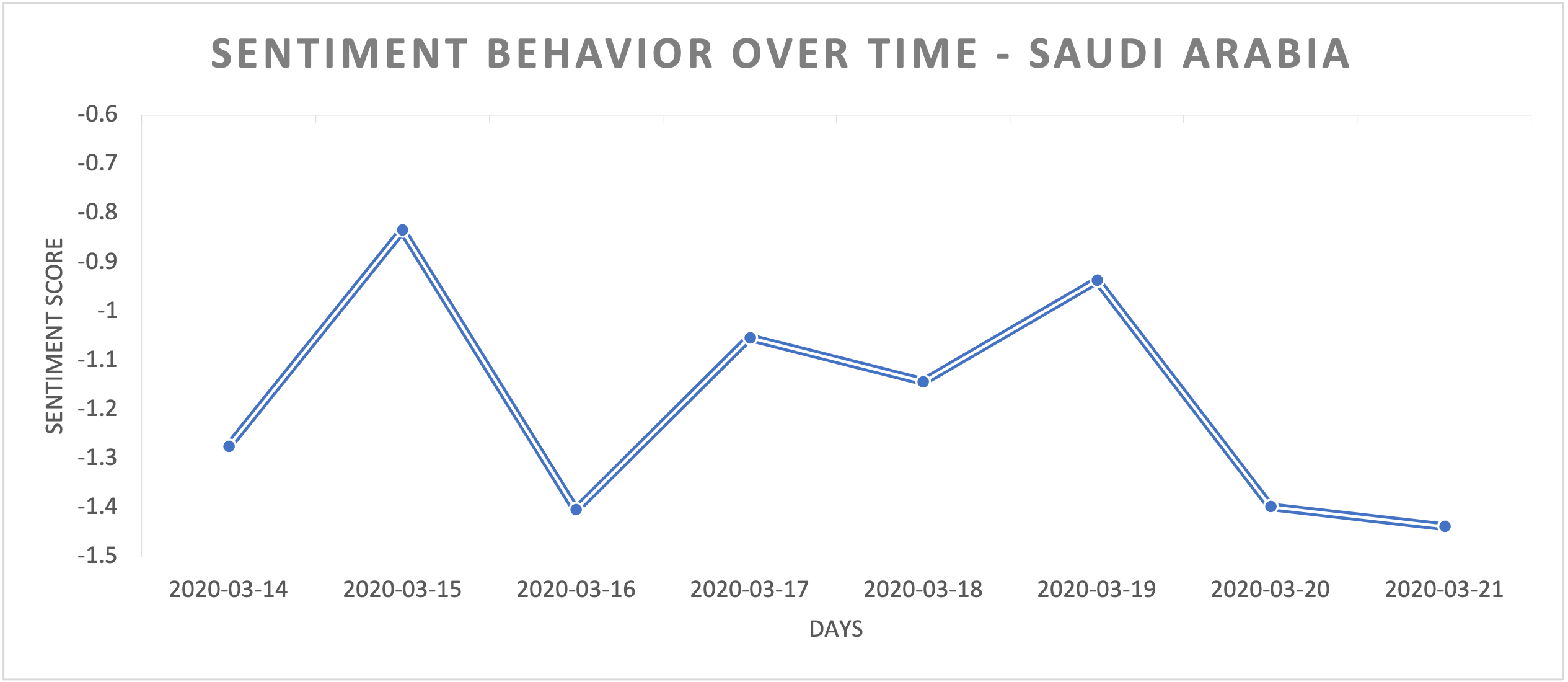}
  \caption{\scriptsize }
 \label{fig:leb-hate-covid}
\end{subfigure}
\begin{subfigure}{.47\textwidth} 
 \centering
  \includegraphics[width=\linewidth]{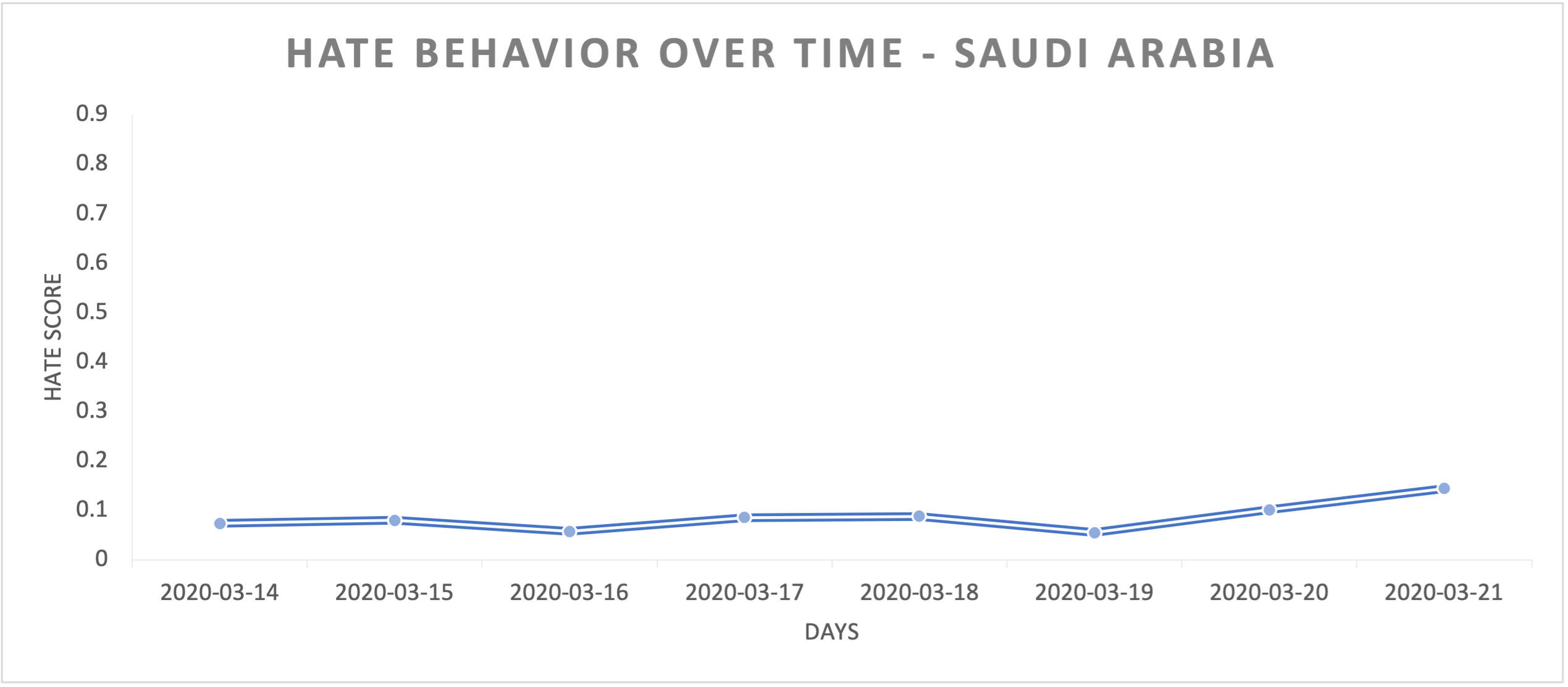}
  \caption{\scriptsize }
  \label{fig:saudi-hate-overtime}
\end{subfigure}
\caption{\small Sentiment behavior over time in Saudi Arabia during COVID-19 Pandemic. The units are days according to the Saudi dataset \cite{saudi-covid19-dataset}.}
\label{fig:saudi-osb-overtime}
\end{figure}

The overall predictions of our localized sentiment classifiers (i.e. Arabic Levantine and Gulf classifiers) on Lebanon and Saudi COVID-19 datasets, show that the sentiment is more negative in Lebanon with 87\% of overall negative vibes than it is in Saudi Arabia with 65\% of negative vibes during COVID-19 pandemic in 2020.  Figure. \ref{fig:saudi-osb-overtime} provides a temporal view of sentiment behavior during the second week of March 2020, when COVID health measures were implemented in Saudi Arabia during COVID-19 pandemic. Overall, the sentiment is shown to decrease over time especially after a positive spike on day 15 of March 2020. The temporal sentiment analysis for Lebanon is not provided in this case study due to the reason that date and time information is not available in the original Lebanon dataset \cite{zahra2020targeted}.

\begin{figure}[h]
\centering
\begin{subfigure}{.47\textwidth}
  \centering
  \includegraphics[width=\linewidth]{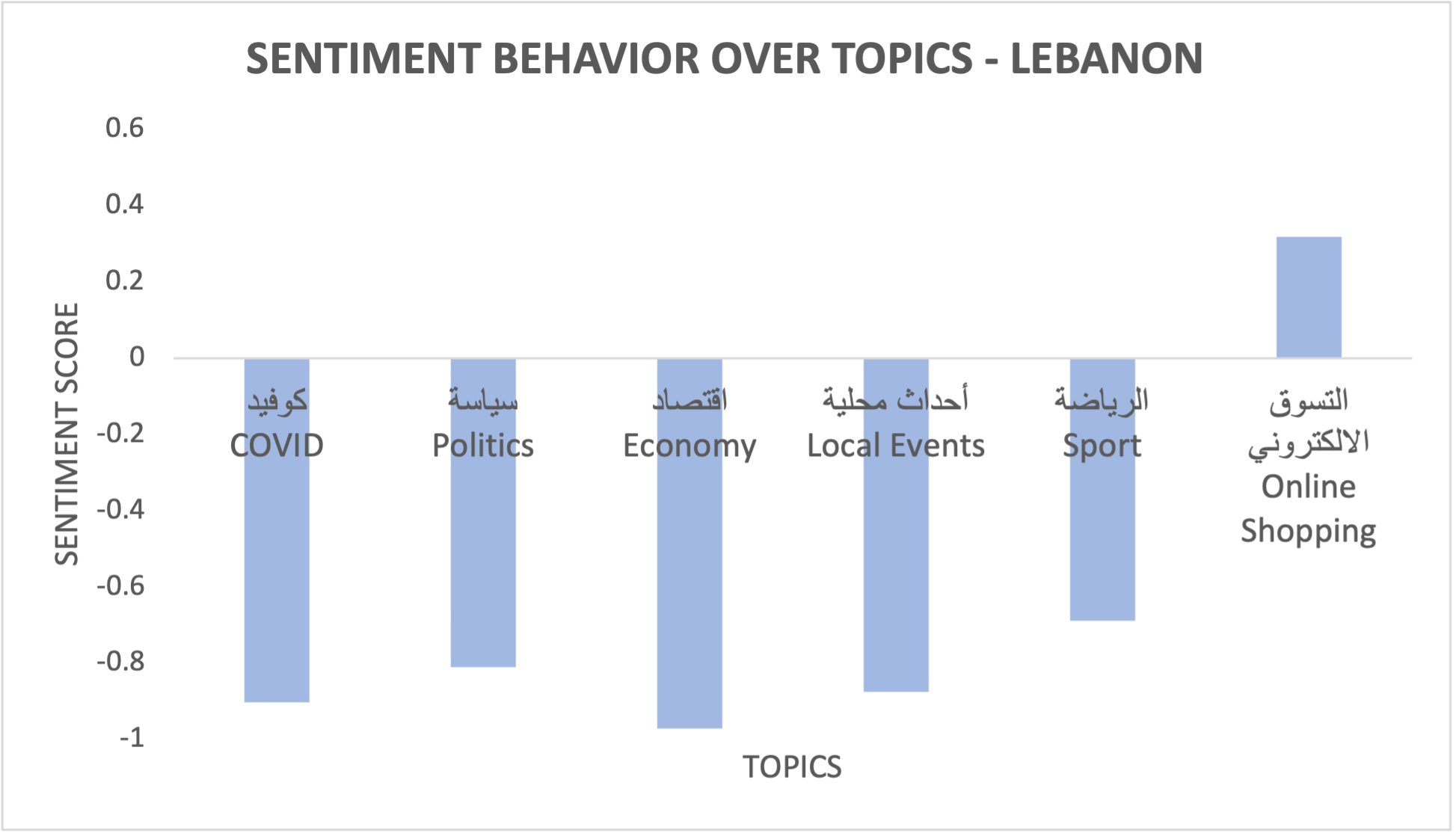}
  \caption{\scriptsize }
  \label{fig:leb-sent-overlltopics}
\end{subfigure}
\begin{subfigure}{.47\textwidth}
  \centering
  \includegraphics[width=\linewidth]{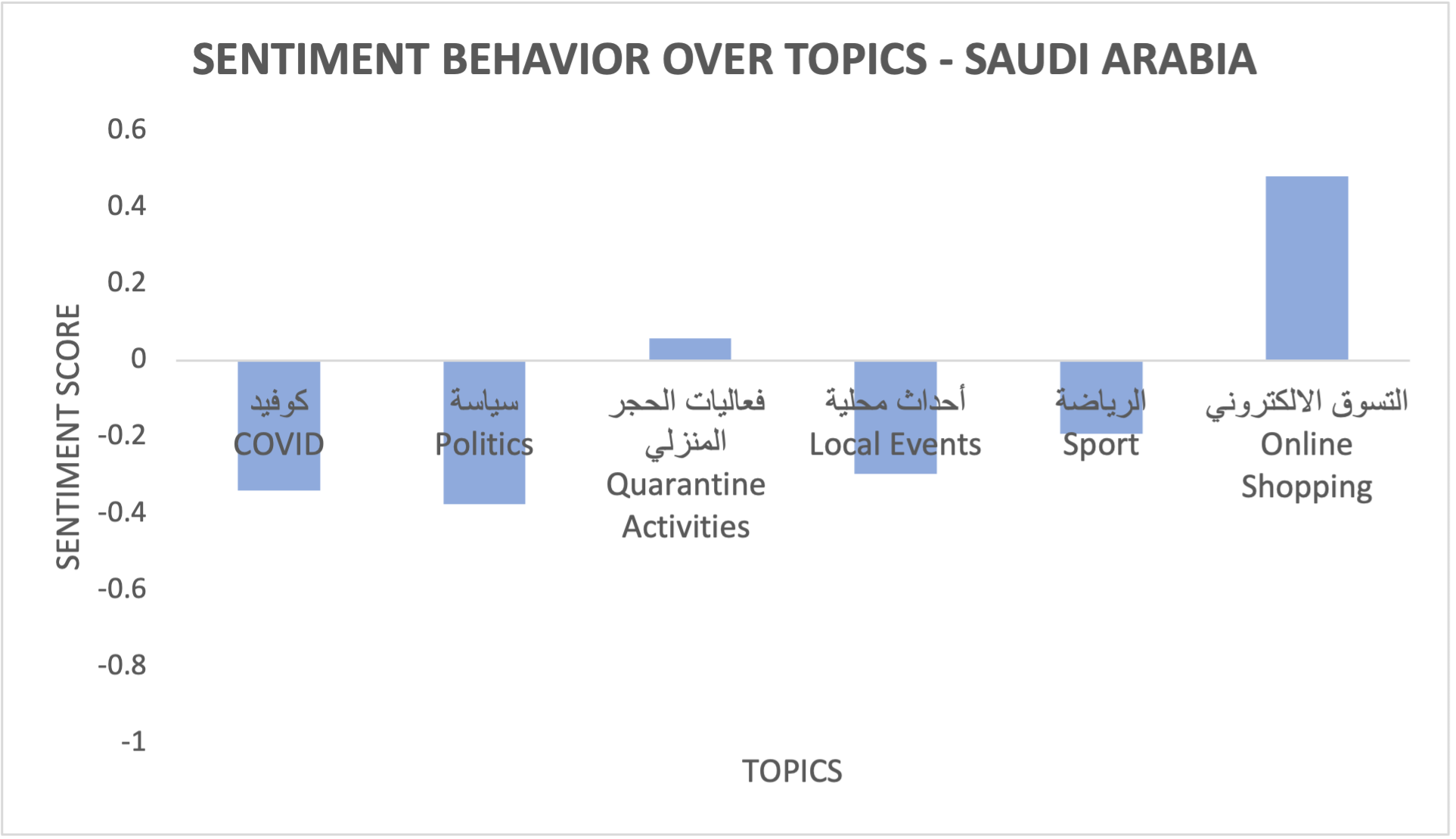}
  \caption{\scriptsize }
  \label{fig:saudi-sent-overlltopics}
\end{subfigure} 
\caption{\small Comparisons of sentiment behavior between Lebanon and Saudi Arabia over inferred topics from COVID-19 data collected from Lebanon and Saudi Arabia during the COVID-19 pandemic.}
\label{fig:leb-saudi-sent-overall-topics}
\end{figure}

To facilitate the understanding of the sentiment behavior in Lebanon and Saudi Arabia during COVID-19 Pandemic,  we provide a deeper analysis of the topics that people were discussing on social media platforms during the pandemic (Figure. \ref{fig:leb-saudi-sent-overall-topics}). Having the topics at hand, we are able to deduce the reasoning of the temporal abstract analysis. In other words, we can understand the reasons and causes of the inferred sentiment behavior. This has helped us to get a clearer picture of the story of events. The topics of both Lebanon and Saudi Arabia show an overall negative sentiment behavior with Lebanon having more negative vibes than Saudi Arabia does; while Saudi Arabia shows a more positive sentiment in two topics (i.e. quarantine/activities and online shopping), Lebanon shows a positive sentiment in one topic (i.e. online shopping) only. Also, the maximum negative score that Saudi sentiment reached is -0.4 compared to the negative peaks of $\approx-1$ that Lebanon sentiment has hit. 

\begin{figure}
\centering
\begin{subfigure}{.485\textwidth} 
  \centering
  \includegraphics[width=.99\linewidth]{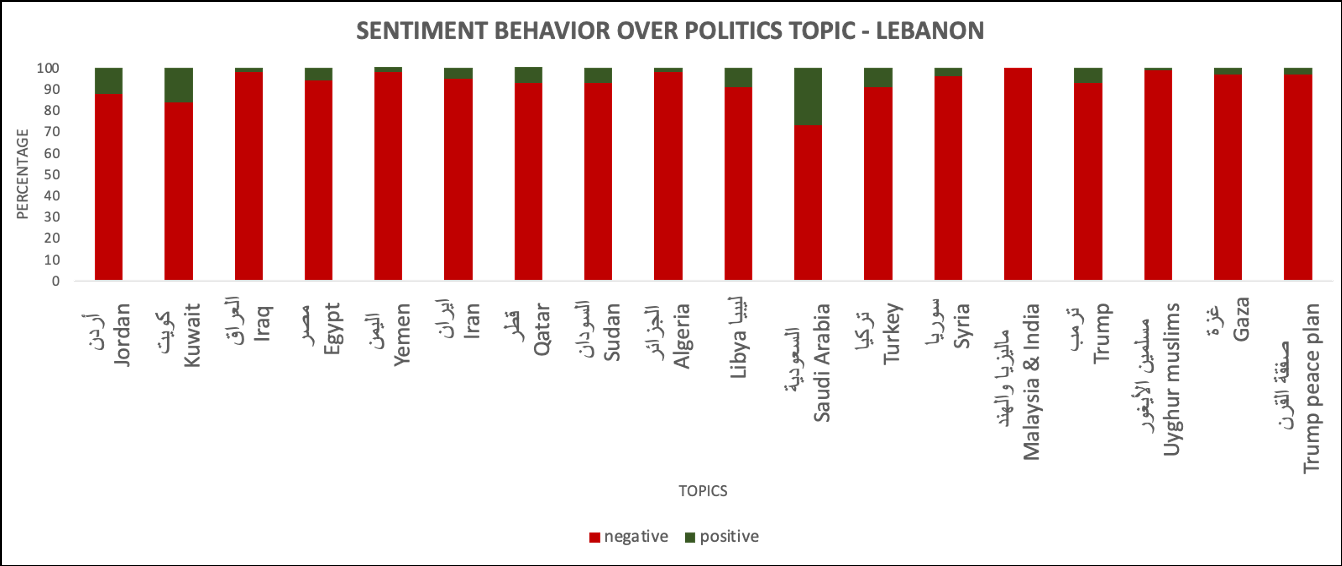}
  \caption{\scriptsize }
  \label{fig:leb-sent-politics}
\end{subfigure}
\begin{subfigure}{.485\textwidth} 
  \centering
  \includegraphics[width=.99\linewidth]{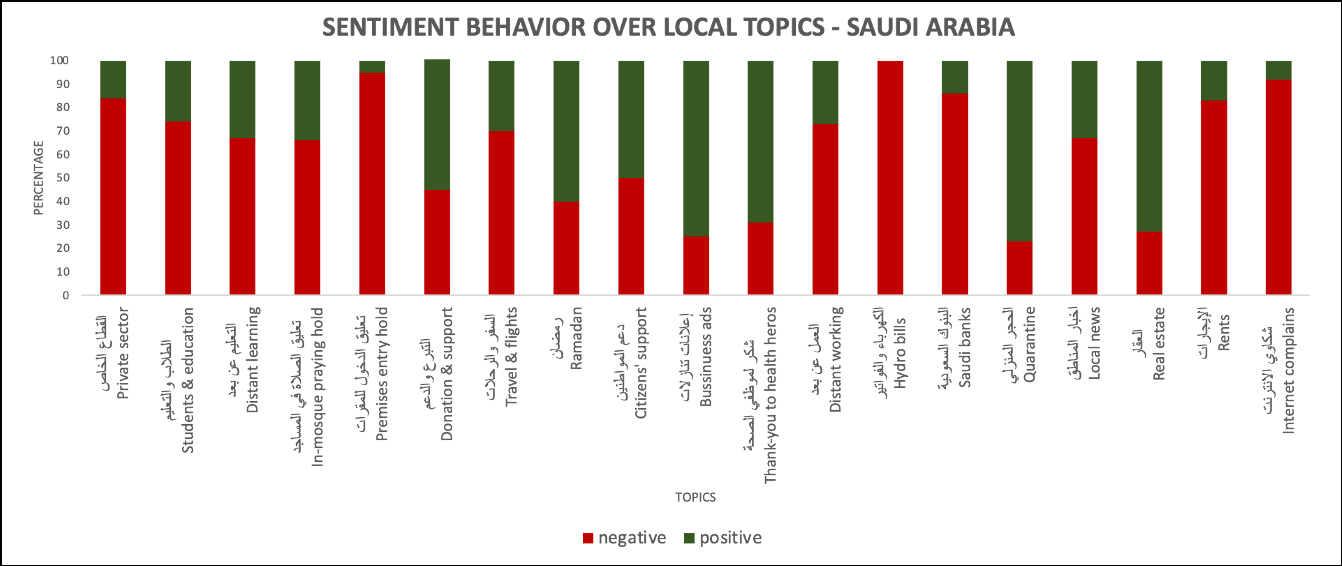}
  \caption{\scriptsize }
  \label{fig:saudi-sent-local}
\end{subfigure}%
\caption{\small Example subtopics of the topics inferred from Lebanon COVID-19 dataset for sentiment behavior analysis in Lebanon and Saudi Arabia.}
\label{fig:leb-saudi-sent-detail-topics}
\end{figure}

 A wider exploratory analysis for the topic groups shown in Figure. \ref{fig:leb-saudi-sent-overall-topics} is illustrated in Figure. \ref{fig:leb-saudi-sent-detail-topics}. The politics topic (Figure. \ref{fig:leb-saudi-sent-overall-topics}) in Lebanon clearly shows the highest negative sentiment; the underlying subtopics (shown in Figure. \ref{fig:leb-sent-politics}) of the politics topic explain the negative behavior by the residents of Lebanon toward international and regional news. It is interesting that the discussions on "Saudi Arabia" followed by "Kuwait" and "Jordan" subtopics, have shown a slight positive sentiment compared to the rest of other politics subtopics. 
Various local Saudi topics are discussed through the subtopics in Figure. \ref{fig:saudi-sent-local}. Subtopics such as "private sector", "students and distant education", "in-mosque praying hold", "hydro and bills", "Saudi banks" and "rents" show mostly negative sentiments. Subtopics that show positive sentiments are related to religion like "Ramadan", and support like "donation", "citizens' support", "thank-you to health workers", and surprisingly "quarantine". The positive sentiment associated with the "quarantine" subtopic reflects that the Saudi citizens seem to be willingly accepting and encouraged to respect the stay-at-home measure in order to prevent the spread of the corona virus. The rest of the topic-based sentiment analysis for both Lebanon and Saudi Arabia can be found in the Appendix section (Figure. \ref{fig:leb-sent-detail-topics-appendix} and \ref{fig:saudi-sent-detail-topics-appendix}).

\subsection{Hate Speech Analysis}
\begin{table}[htbp]
  \centering
  \caption{\small The performance of our localized Arabic-Levantine and Arabic-Gulf hate classifiers on an external hate speech dataset in Levantine dialect, in terms of accuracy, precision, and recall.}
  \resizebox{.7\textwidth}{!}{%
    \begin{tabular}{c|p{3.915em}|c|c|c|c|c|c|c}
    \toprule
    \multirow{5}[10]{*}{\begin{sideways}\textbf{Model}\end{sideways}} & \multicolumn{8}{c}{\textbf{En$->$Ar-Levantine evaluated on Let-Mi dataset \cite{mulki2021let}}} \\
\cmidrule{2-9}          & \multicolumn{1}{c|}{} & \multicolumn{3}{c|}{\textbf{Hate}} & \multicolumn{3}{c|}{\textbf{No-Hate}} &  \\
\cmidrule{2-9}          & \multicolumn{1}{c|}{} & \textbf{Precision} & \textbf{Recall} & \textbf{F-Score} & \textbf{Precision} & \textbf{Recall} & \textbf{F-Score} & \textbf{Accuracy} \\
\cmidrule{2-9}          & En$->$Ar-Levantine & 0.76  & 0.67  & 0.71  & 0.71  & 0.79  & 0.75  & 73 \\
\cmidrule{2-9}          &  En$->$Ar-Gulf & 0.81  & 0.38  & 0.51  & 0.6   & 0.91  & 0.72  & 65 \\
    \bottomrule
    \end{tabular}
    }
  \label{tab:mlmd-hate-osb-external-en}
\end{table}
The content-localization based hate classifiers are shown to have been able to efficiently learn representative features from the localized hate datasets to detect the hate content correctly using the validation split data, as seen in Table. \ref{tab:mlmd-hate-sent-val}. According to the results, our localized Levant and Gulf hate classifiers could accurately classify hate and non-hate content at validation f-score between 0.68\% and 0.70\% for Levant and Gulf hate classifiers, respectively. Further assessments on an external native Levant hate dataset have illustrated the validity of our content-localization approach; the Levant hate classifier could successfully and reliably recall 67\% of hate content from the native Levant hate tweets (i.e. from Let-Mi datasets \cite{mulki2021let}) while maintain a high precision performance of 76\% for the hate class. On the other hands, the Gulf hate classifier could only recall 38\% of the hate presence from the same set of Levant hate tweets at a 81\% of hate precision.
 
We list example sentences expressed in Arabic-Levantine dialect (i.e. taken from Let-Mi datasets \cite{mulki2021let}), which the localized Gulf hate classifier has classified as non-hate, while our localized Levantine classifier has been able to classify as hate content: (1) {\scriptsize "\< اي نحن ما منقبلها صرماية بإجرنا ...مبروك. ع راسكم>}, (2) {\scriptsize "\<انشالله بيقبر قلبك عن قريب ...يافهيمة عصرك>"}, (3) {\scriptsize "\<ضبي لسانك احسنلك>}, (4) {\scriptsize "\<انشالله بيقبرك إنتي وعيلتك>"}.

As seen in the examples listed above, the same language has got different localized dialects; an expression in a certain dialect means something else in another and it is used in a different context. Ignoring such a feature can negatively impact the learning models so much that they end up generating misleading outputs. For instance, the Levantine idiom "{\scriptsize \<صرماية بإجرنا>}" -which means "shoes in our foot"- is a very local expression that Levantine people use in a negative situation (i.e. usually when in anger and it is used for swearing); however, the Gulf people do not use this idiom with the same structure; its equivalence though can be "{\scriptsize \<شبشب في رجلي>" or "\<نعال في رجلي>}". The same applies to the toxic expressions "{\scriptsize \<بيقبرك, بيقبر قلبك, ضبي لسانك>}" - literally translated as "bury you, bury your heart, hold or fold your tongue"- are used exclusively by Levantine people to express anger or dissatisfaction. 

This finding highlights the importance of distinguishing dialects of the very same language and their localized contextual meanings. Overlooking those differences results in inaccurate understanding of the target dialect, which in turn leads to misleading and imprecise analysis of online social behaviors. 

\begin{figure*}[h]
\centering
\begin{subfigure}{.47\textwidth}
  \centering
  \includegraphics[width=\linewidth]{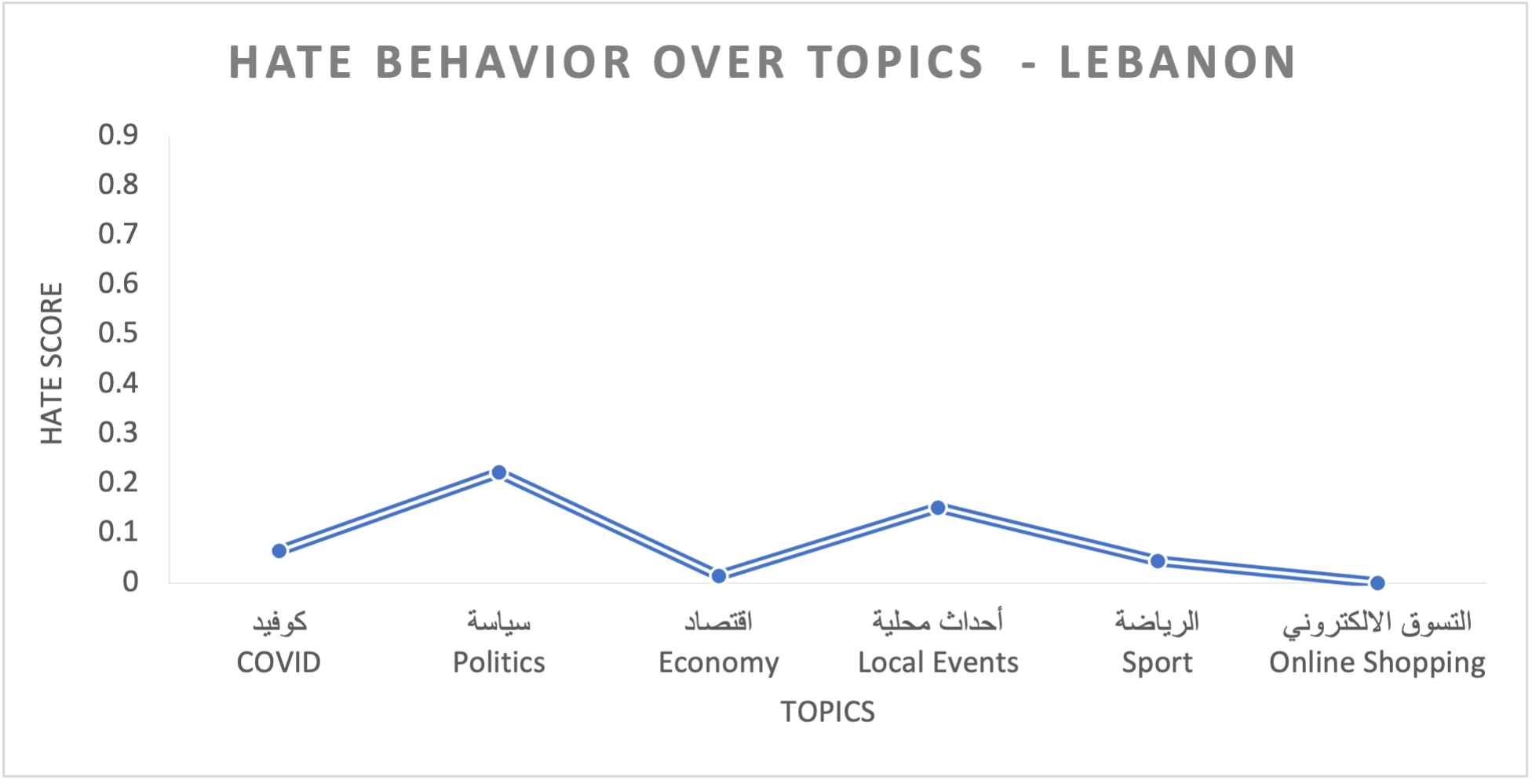}
  \caption{\scriptsize }
  \label{fig:leb-hate-overlltopics}
\end{subfigure}
\begin{subfigure}{.47\textwidth}
  \centering
  \includegraphics[width=\linewidth]{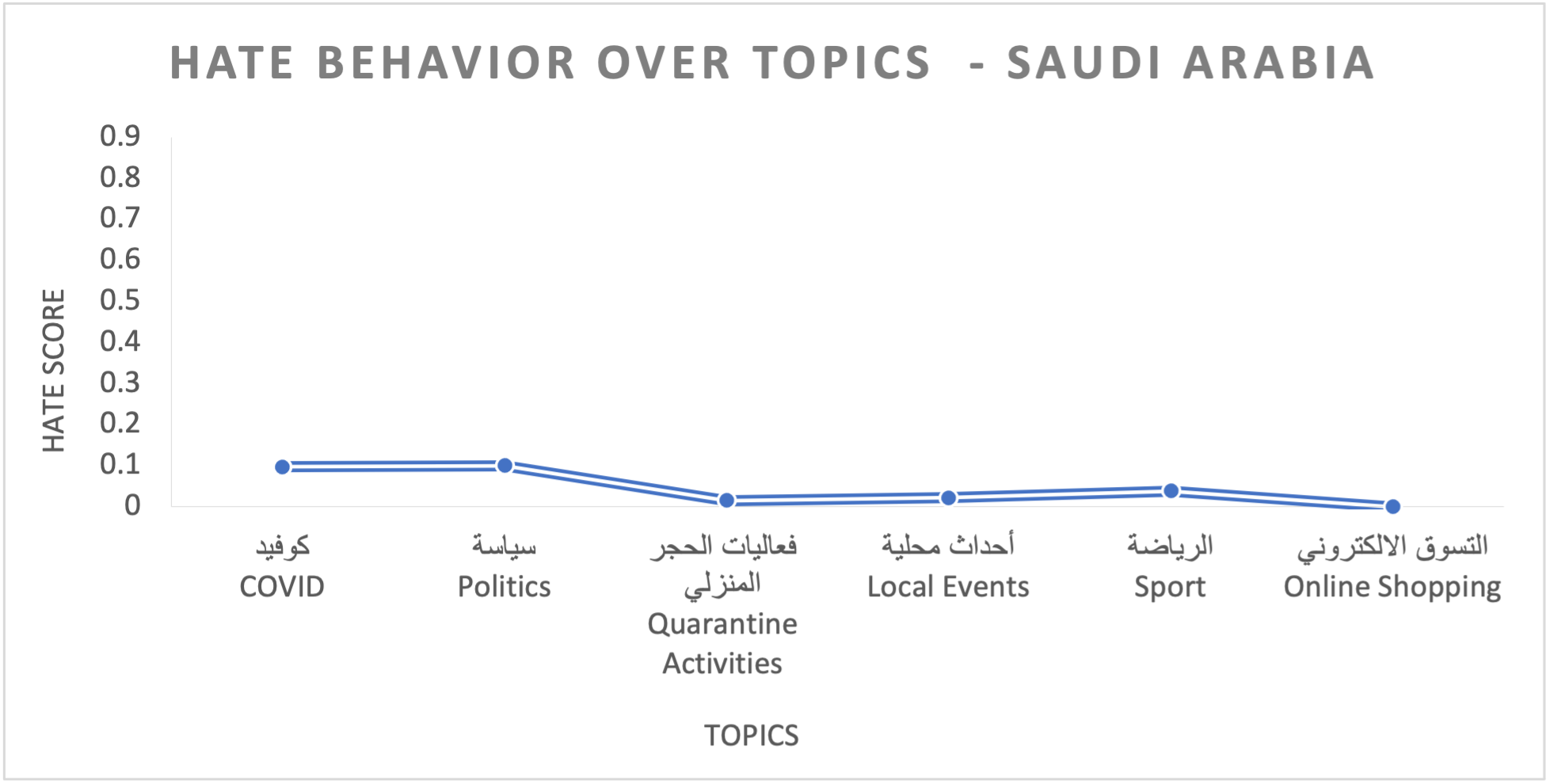}
  \caption{\scriptsize }
  \label{fig:saudi-hate-overlltopics}
\end{subfigure}
\caption{\small Comparisons of hate behavior between Lebanon and Saudi Arabia over inferred topics from COVID-19 data collected from Lebanon and Saudi Arabia during the COVID-19 pandemic.}
\label{fig:leb-saudi-hate-overall-topics}
\end{figure*}
\begin{figure*}[h]
\centering
\begin{subfigure}{.485\textwidth} 
  \centering
  \includegraphics[width=.99\linewidth]{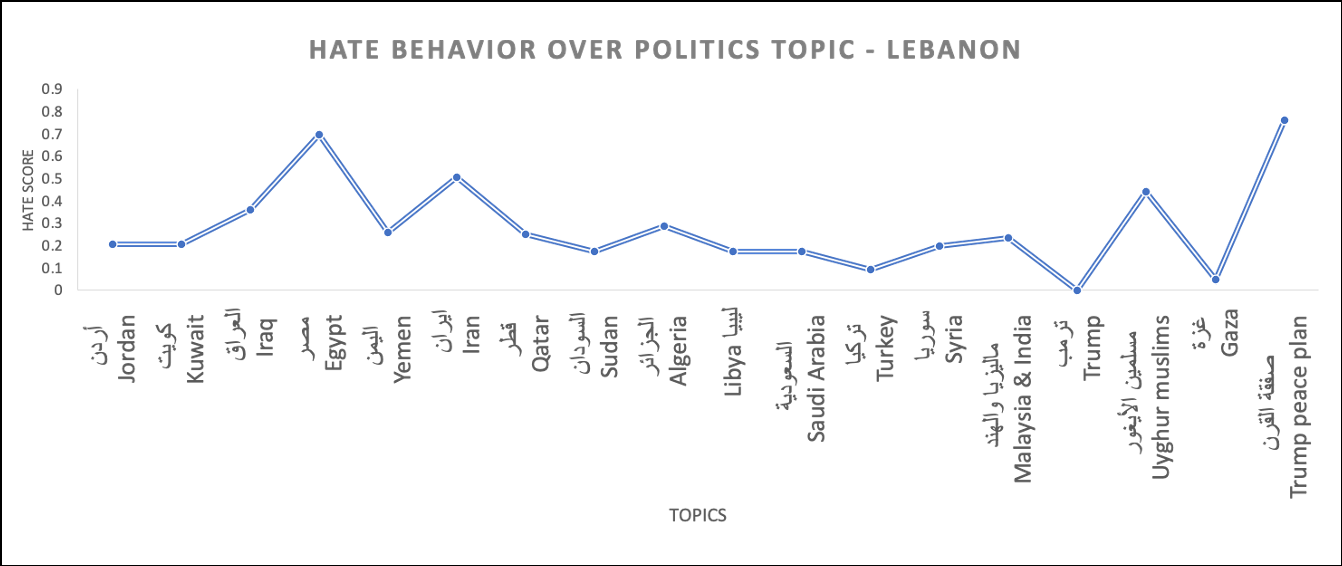}
  \caption{\scriptsize }
  \label{fig:leb-hate-politics}
\end{subfigure}
\begin{subfigure}{.485\textwidth} 
  \centering
  \includegraphics[width=.99\linewidth]{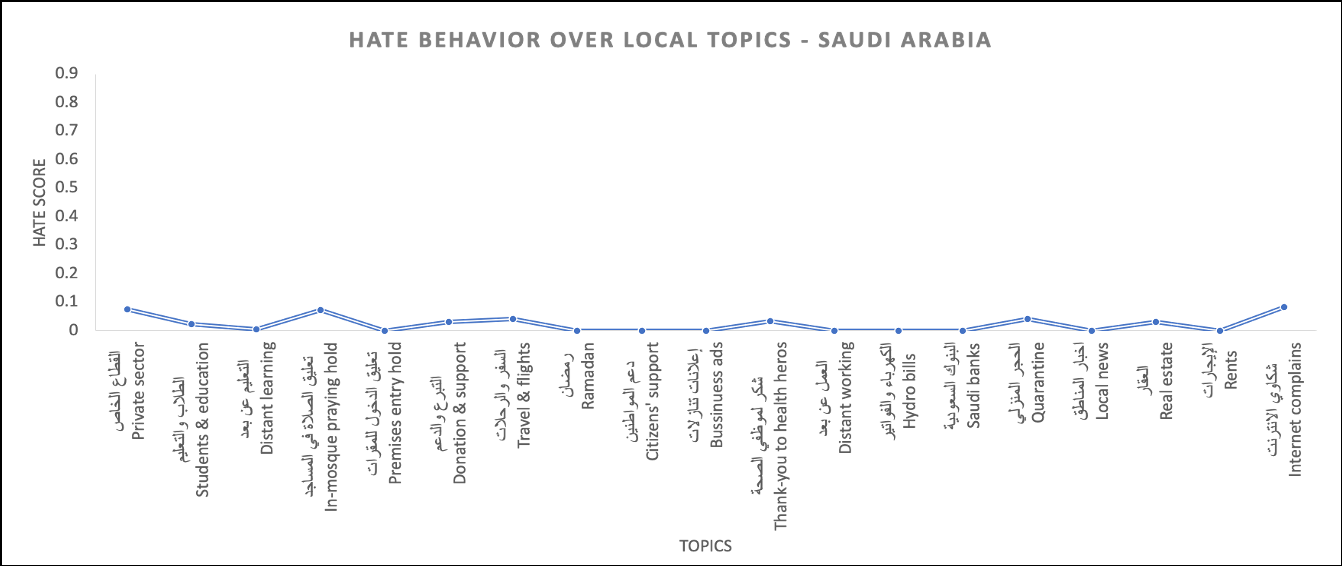}
  \caption{\scriptsize }
  \label{fig:saudi-hate-local}
\end{subfigure}%
\caption{\small Example subtopics of the topics inferred from Lebanon COVID-19 dataset for hate behavior analysis in Lebanon and Saudi Arabia.}
\label{fig:leb-hate-detail-topics}
\end{figure*}
We have utilized our localized Arabic Levantine and Gulf hate classifiers to analyze the hate speech behavior during COVID-19 pandemic in Lebanon and Saudi Arabia. Our results show the overall hate behavior detected in Lebanon is $> 2x$ greater than (18\%) the hate behavior detected in Saudi Arabia (7\%). This is clear in the hate behavior depicted in Figure. \ref{fig:leb-saudi-hate-overall-topics}; hate behavior in Lebanon is higher than that in Saudi Arabia during COVID-19 in 2020, especially in topics related to politics and local events, where the hate behavior scores the highest (Figure. \ref{fig:leb-hate-overlltopics}). Saudi Arabia has shown a slight hate behavior only in topics related to COVID-19 and politics; however, the detected hate behavior in Saudi Arabia is still $\approx 2x$ lower than that in Lebanon (Figure. \ref{fig:saudi-hate-overlltopics}). COVID topic in Saudi Arabia (Figure. \ref{fig:saudi-hate-overlltopics}) is slightly higher in hate score compared to its corresponding in Lebanon (Figure. \ref{fig:leb-hate-overlltopics}); this indicates that the people in Saudi Arabia seem to have been more upset about the virus spread and its consequences. "Sport" topic in both Lebanon and Saudi Arabia shows an insignificant level of hate behavior, while topics "economy", "quarantine activities", and "online shopping" score zero hate behavior in both Lebanon and Saudi Arabia.

Politics topic in Lebanon, which has the highest hate score during the COVID-19 pandemic in Lebanon (Figure. \ref{fig:leb-hate-overlltopics}), discusses eighteen subtopics (Figure. \ref{fig:leb-hate-politics}), out of which are "Iran", "Egypt", "Uyghur Muslims", and "Trump peace plan". Those subtopics show the highest hate behavior in the politics topic. Throughout the local topics discussed in Saudi COVID-19 data (Figure. \ref{fig:saudi-hate-local}), subtopics "private sector", "in-mosque praying hold", and "internet complaints" show a slight amount of hate behavior. The hate behavior detected in "private sector" reflects peoples' complaints about private sector being late in implementing restriction measures during the pandemic in Saudi Arabia. Religion is an essential part of life in Saudi Arabia; people go to mosques five times a day to perform five prayers every day. Therefore, implementing the measure of closing down mosques during the pandemic in Saudi Arabia was a major frustration to many, and that might explain the slight presence of hate behavior in the associated subtopic. The hate behavior detected in "internet complaints" subtopic is expected; during the pandemic all schools, universities, and work shifted to online, and people were instructed to stay home; however, some suffered from internet connection issues, which is most likely the cause of hate behavior.
The rest of the topic-based hate analysis for both Lebanon and Saudi Arabia can be found in the Appendix section (Figure. \ref{fig:leb-hate-detail-topics-appendix} and \ref{fig:saudi-hate-detail-topics-appendix}).

\section{Conclusion}
\label{conc}
This paper addresses the issue of sub-optimal performance of existing machine translation systems in accurately translating informal messages of low-resourced languages like dialectal Arabic, and proposes a system that utilizes content-localization based neural machine translation (NMT) models customized for informal communications on social media. This being said, these NMT models do not only localize the context and culture of informal messages into different dialects of a language, but also they localize the OSN culture into the localization of these messages. We localize the content of sentiment and hate speech datasets from English language to two low-resourced Arabic dialects (i.e. Levantine and Gulf) and develop four OSB (i.e. sentiment and hate speech behaviors in this paper) classifiers for both Levantine and Gulf dialects using the localized data resources. We then evaluate the performance of our proposed OSB classifiers using two approaches: (1) using external data resources collected and manually annotated in native Levanitne and Gulf dialects, (2) using proof-of-concept case study on COVID-19 data collected during the pandemic in 2020 from Lebanon (Levant dialect) and Saudi Arabia (Gulf dialect) regions. Both evaluation approaches have proven the efficacy of our proposed system in learning sentiment and hate speech from localized data resources; this has been shown in the high performance results in terms of precision, recall, f-score, and accuracy, for both sentiment and hate speech tasks in Levantine and Gulf dialects. Further, our experimental results shed light on the importance of considering different dialects within the very same language to ensure effective and accurate OSB analysis. This can be seen in our localized Levant hate model being able to detect hate content from native Levant messages whereas our localized Gulf hate model has shown low performance to do so. This proves that by overlooking dialectal aspects within a language, we can miss out on valuable insights and perspectives which in turn would lead to inaccurate and misleading results and hence improper policy and decision making. The causing and reasoning behind predicted online sentiment and hate behaviors are achieved through our proposed unsupervised learning methodology for data exploration and interpretation. We adopt topic modeling and phrase extraction approach for capturing insightful topics, trends, and concerns formed during the pandemic and automatically providing coherent interpretations to the inferred topics regardless of the language/dialect used and without human intervention. We opt to use unsupervised learning techniques that omit the requirements of language-dependent preprocessing tools to remove noises and retain informative pieces of given data. Our BERTopic models have shown a robust performance against data noises and successfully identified meaningful topics from both Lebanon and Saudi Arabia COVID-19 datasets representing two different Arabic dialects: Levantine and Gulf. To coherently interpret the inferred topics by our topic models, RAKE algorithm has illustrated superior language/dialect independence capability to automatically and dynamically extract the most representative phrases that best describe each topic better than the interpretation of single keywords alone. 

For future directions, we plan to extend our system to include more of the Arabic dialects like Yemeni,  Iraqi, and Egyptian. We also plan to expand ours OSB modeling to various social behaviors including sarcasm, emotions, and optimism/pessimism. Moreover, we are interested in dynamic topic modeling to automatically track changes of topics over time. This will serve as an assisting tool to understand formed trends and concerns in an event through their evolution over time; this will allow us to improve the analysis of online social behavior of citizens in smart cities.

\appendix
\section*{Appendix}
\begin{figure}[h]
\centering
\begin{subfigure}{.5\textwidth} 
  \centering
  \includegraphics[width=.99\linewidth]{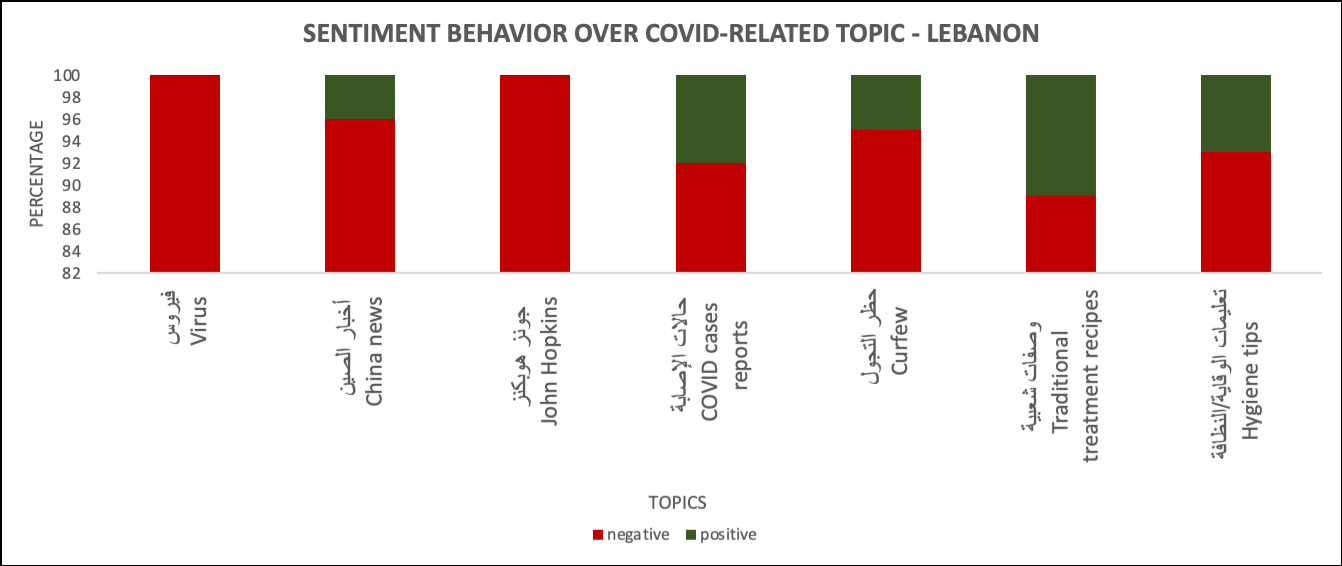}
  \caption{\scriptsize }
  \label{fig:leb-sent-covid-appendix}
\end{subfigure}%
\begin{subfigure}{.5\textwidth} 
  \centering
  \includegraphics[width=.99\linewidth]{images/leb-topics-sent-hate/leb-sent-politics2.png}
  \caption{\scriptsize }
  \label{fig:leb-sent-politics-appendix}
\end{subfigure}
\begin{subfigure}{.5\textwidth} 
  \centering
  \includegraphics[width=.99\linewidth]{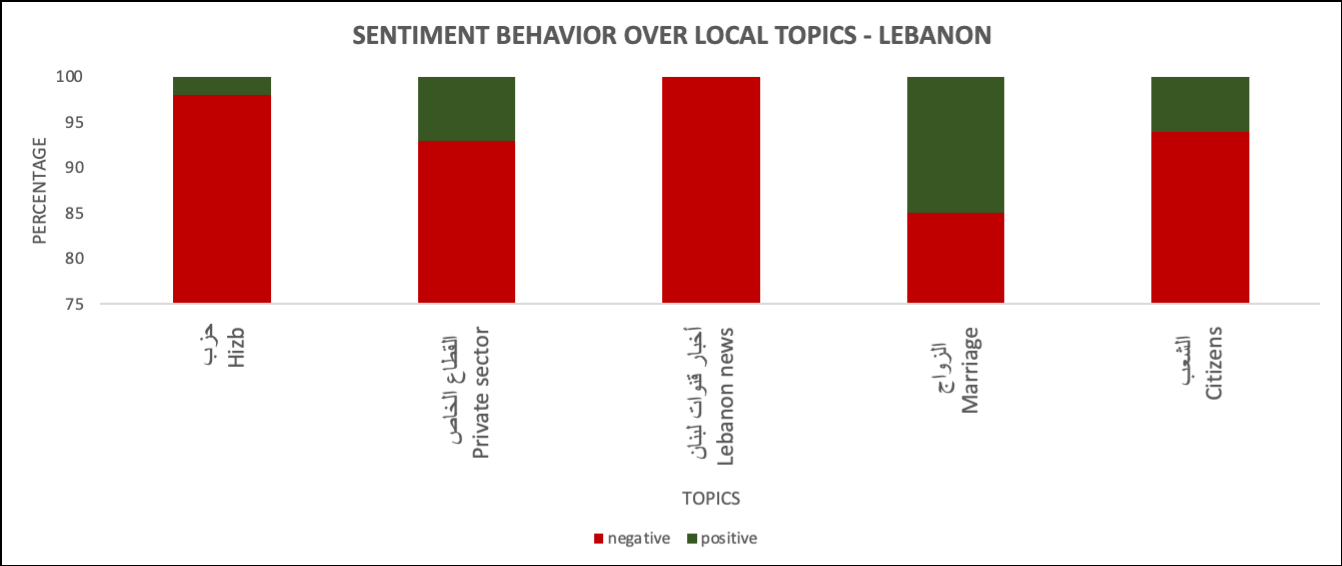}
  \caption{\scriptsize }
  \label{fig:leb-sent-local-appendix}
\end{subfigure}%
\begin{subfigure}{.5\textwidth} 
  \centering
  \includegraphics[width=.99\linewidth]{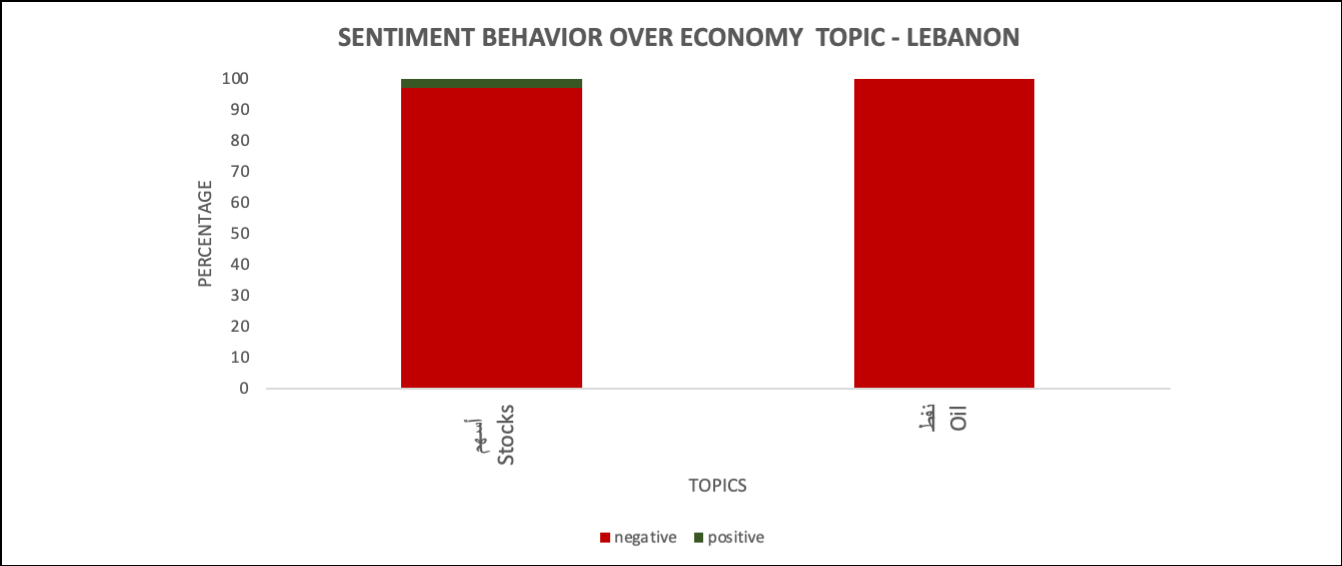}
  \caption{\scriptsize }
  \label{fig:leb-sent-economy-appendix}
\end{subfigure}
\begin{subfigure}{.5\textwidth} 
  \centering
  \includegraphics[width=.99\linewidth]{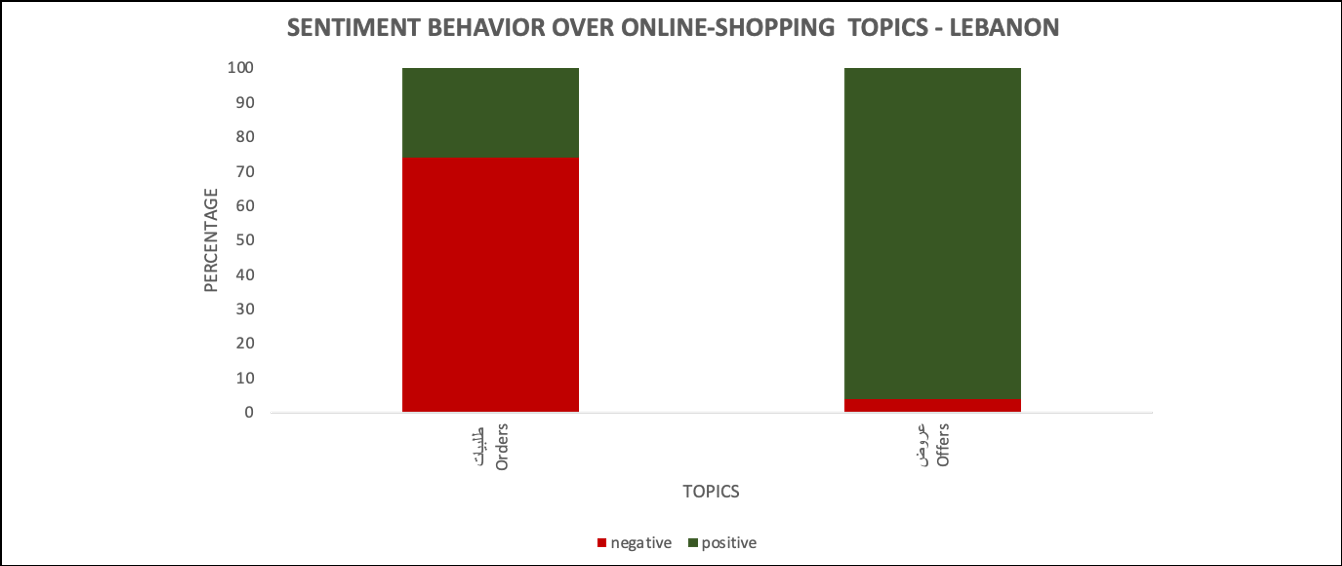}
  \caption{\scriptsize }
  \label{fig:leb-sent-shopping-appendix}
\end{subfigure}%
\begin{subfigure}{.5\textwidth} 
  \centering
  \includegraphics[width=.99\linewidth]{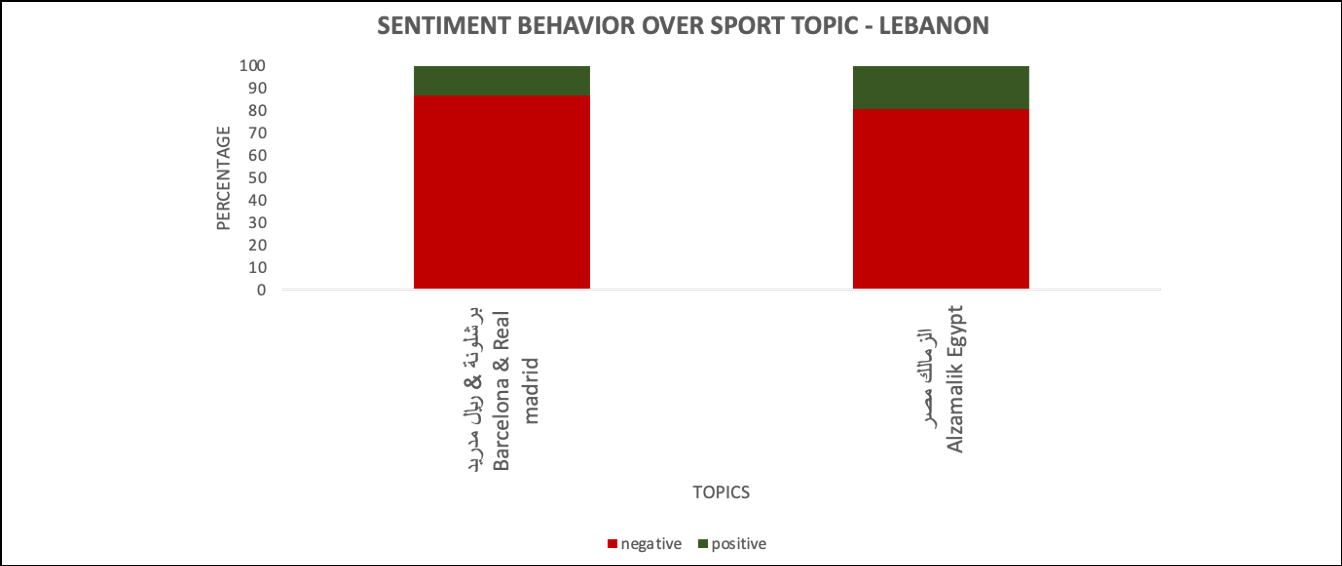}
  \caption{\scriptsize }
  \label{fig:leb-sent-sport-appendix}
\end{subfigure}
\caption{\small Subtopics of the topics inferred from Lebanon COVID-19 dataset for sentiment behavior analysis in Lebanon.}
\label{fig:leb-sent-detail-topics-appendix}
\end{figure}


\begin{figure}
\centering
\begin{subfigure}{.5\textwidth} 
  \centering
  \includegraphics[width=.99\linewidth]{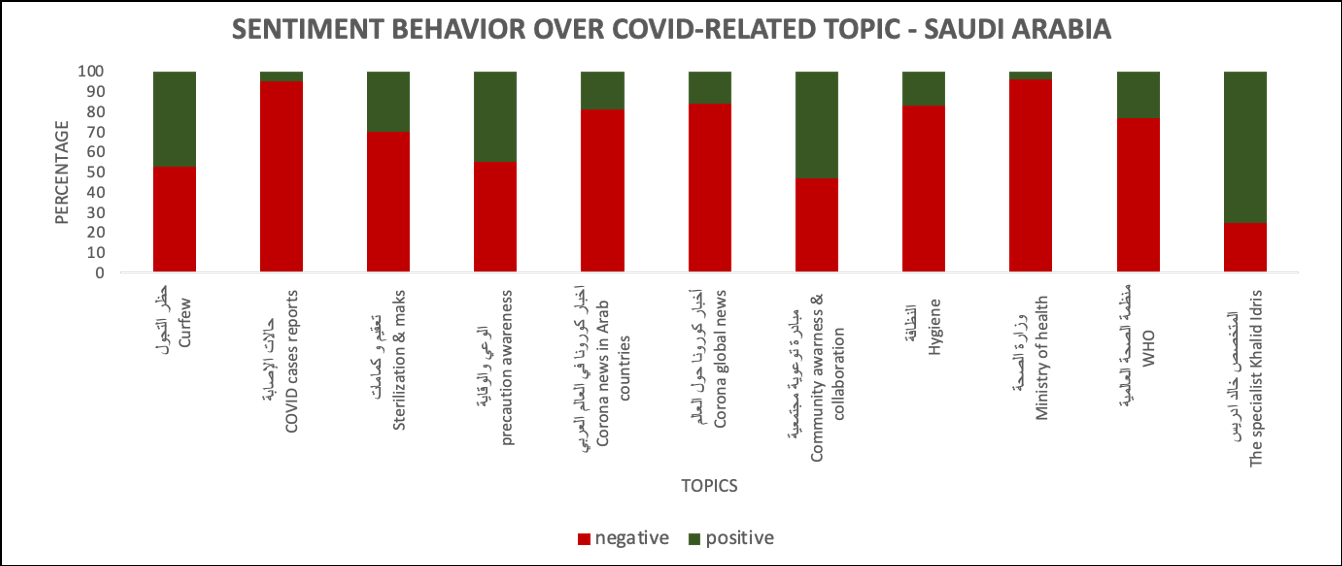}
  \caption{\scriptsize }
  \label{fig:saudi-sent-covid-appendix}
\end{subfigure}%
\begin{subfigure}{.5\textwidth} 
  \centering
  \includegraphics[width=.99\linewidth]{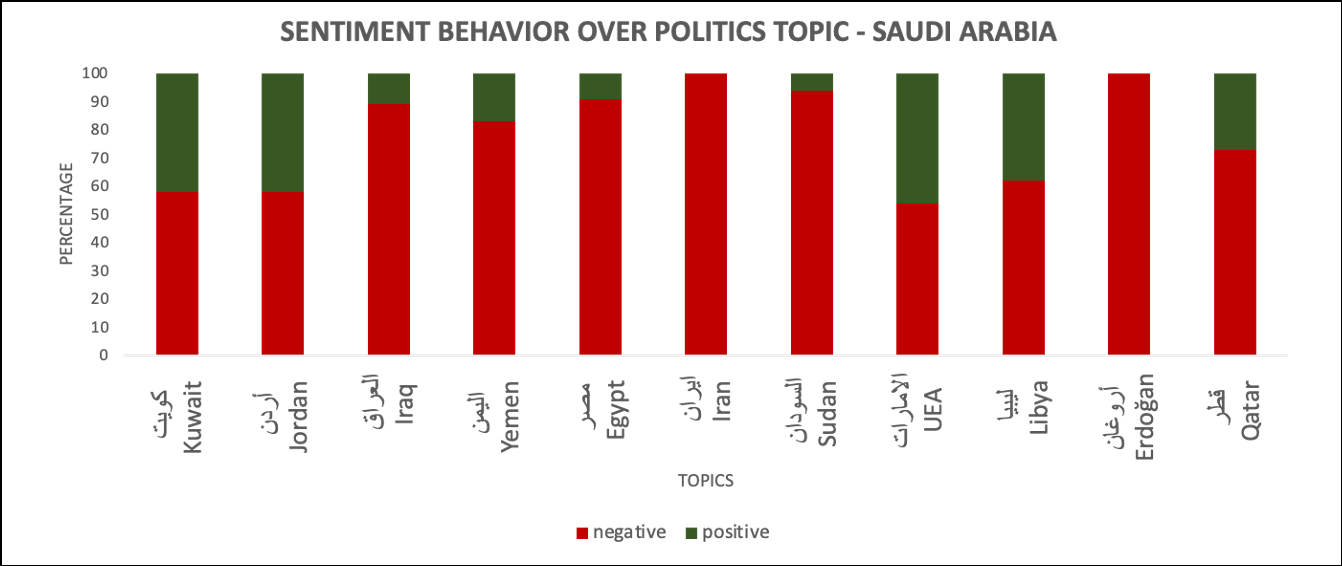}
  \caption{\scriptsize }
  \label{fig:saudi-sent-politics-appendix}
\end{subfigure}
\begin{subfigure}{.5\textwidth} 
  \centering
  \includegraphics[width=.99\linewidth]{images/saudi-sent-hate-topics/saudi-sent-local2.png}
  \caption{\scriptsize }
  \label{fig:saudi-sent-local-appendix}
\end{subfigure}%
\begin{subfigure}{.5\textwidth} 
  \centering
  \includegraphics[width=.99\linewidth]{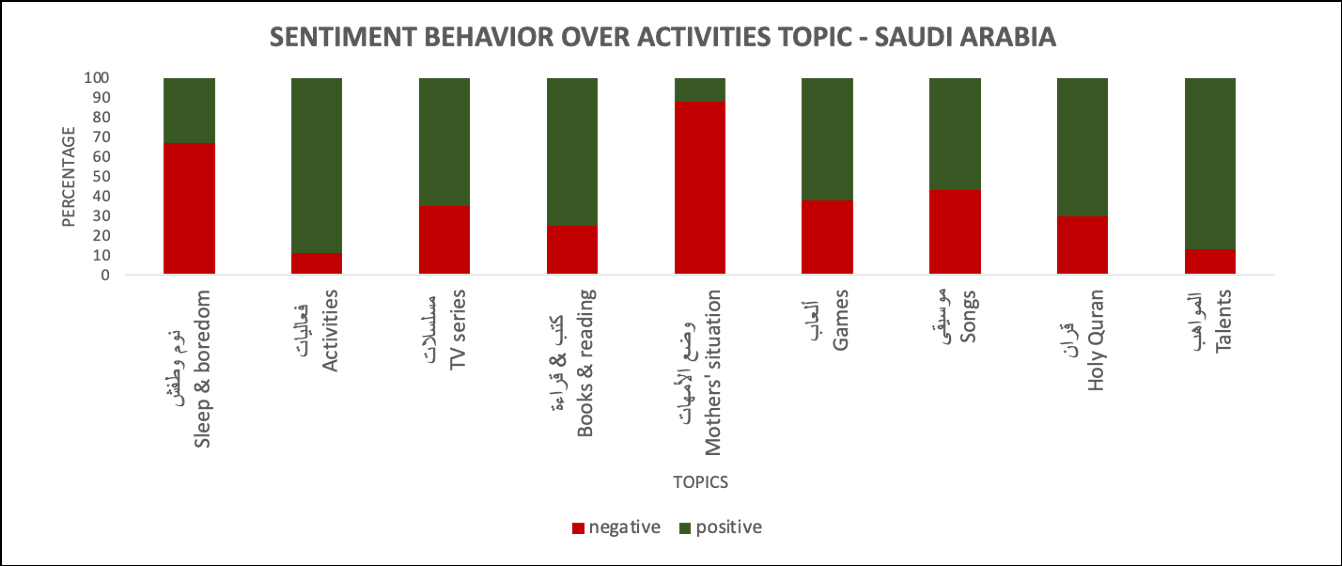}
  \caption{\scriptsize }
  \label{fig:saudi-sent-activities-appendix}
\end{subfigure}
\begin{subfigure}{.5\textwidth} 
  \centering
  \includegraphics[width=.99\linewidth]{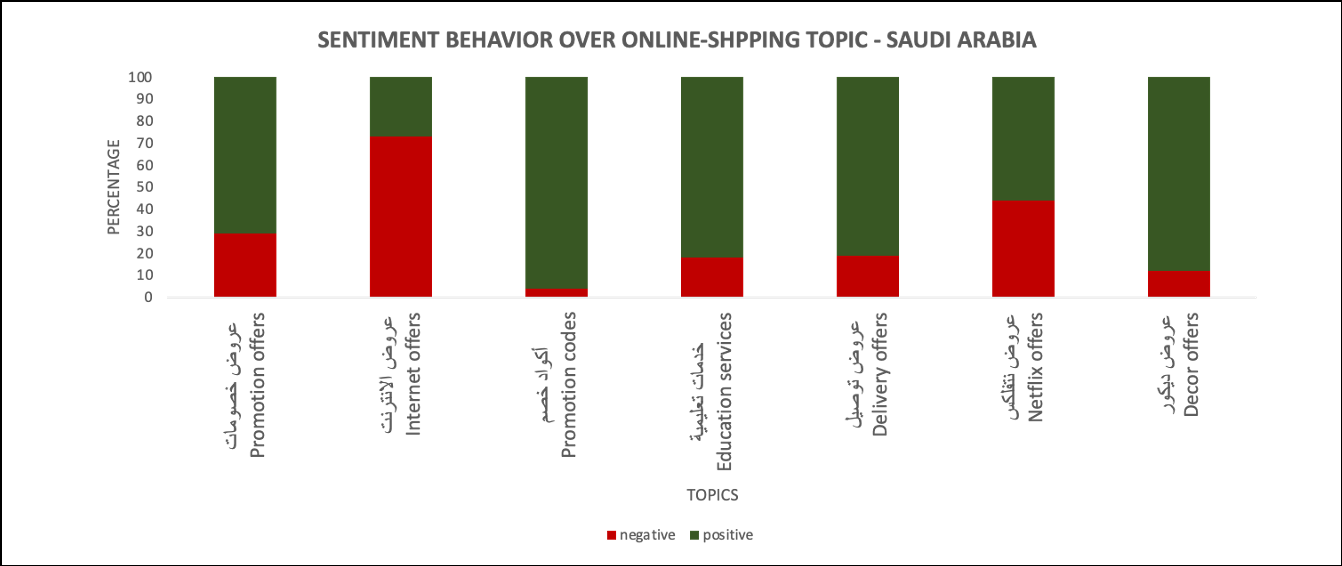}
  \caption{\scriptsize }
  \label{fig:saudi-sent-shopping-appendix}
\end{subfigure}%
\begin{subfigure}{.5\textwidth} 
  \centering
  \includegraphics[width=.99\linewidth]{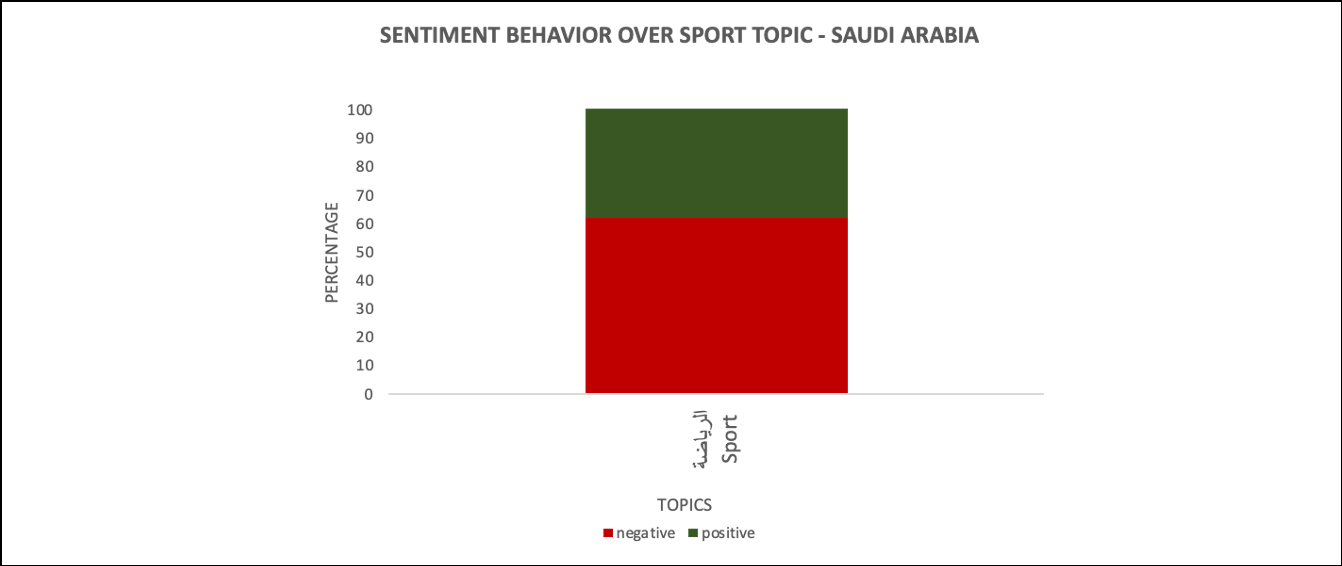}
  \caption{\scriptsize }
  \label{fig:saudi-sent-sport-appendix}
\end{subfigure}
\caption{\small Subtopics of the topics inferred from Saudi Arabia COVID-19 dataset for sentiment behavior analysis in Saudi Arabia.}
\label{fig:saudi-sent-detail-topics-appendix}
\end{figure}

\begin{figure}[h]
\centering
\begin{subfigure}{.5\textwidth} 
  \centering
  \includegraphics[width=.99\linewidth]{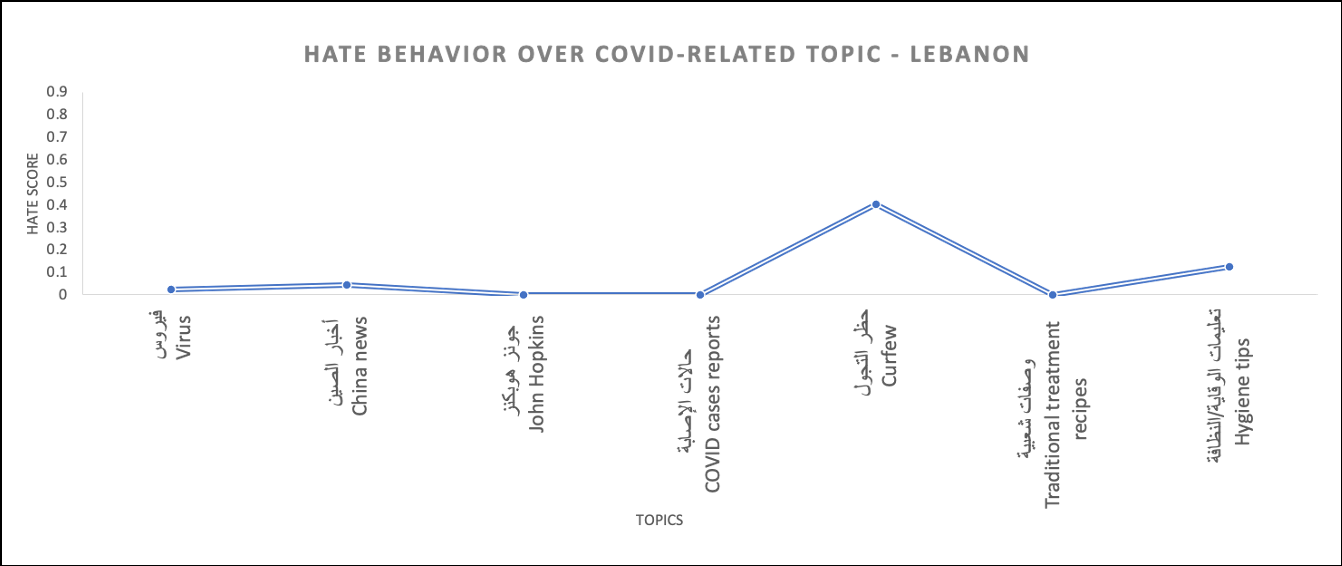}
  \caption{\scriptsize }
  \label{fig:leb-hate-covid-appendix}
\end{subfigure}%
\begin{subfigure}{.5\textwidth} 
  \centering
  \includegraphics[width=.99\linewidth]{images/leb-topics-sent-hate/leb-hate-politics2.png}
  \caption{\scriptsize }
  \label{fig:leb-hate-politics-appendix}
\end{subfigure}
\begin{subfigure}{.5\textwidth} 
  \centering
  \includegraphics[width=.99\linewidth]{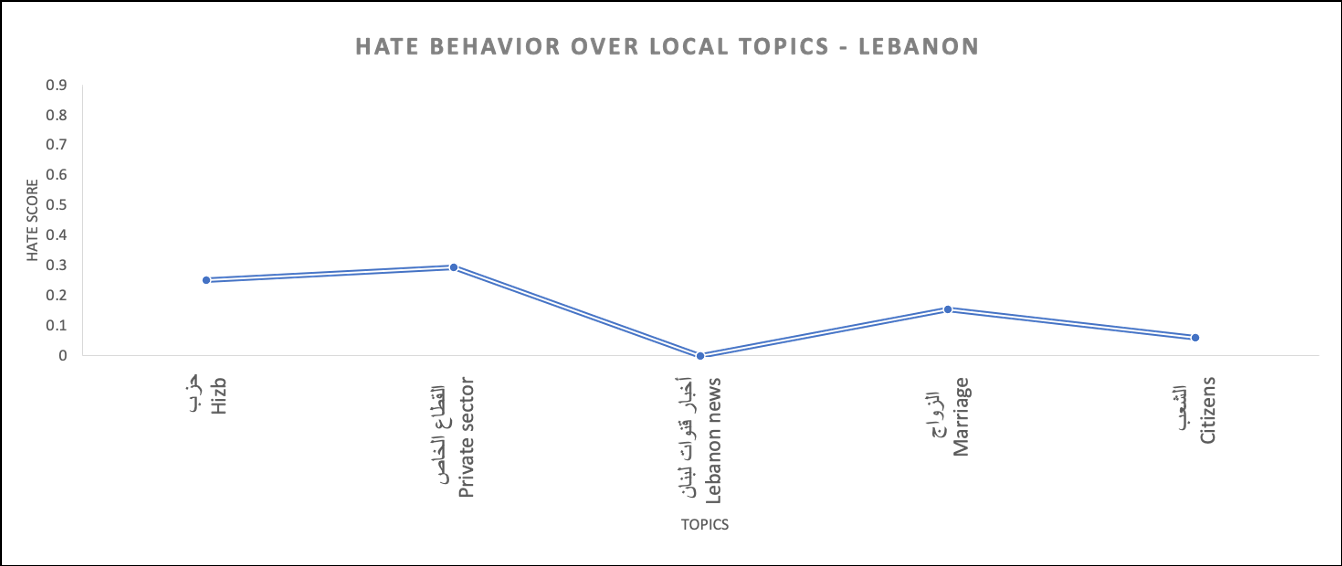}
  \caption{\scriptsize }
  \label{fig:leb-hate-local-appendix}
\end{subfigure}%
\begin{subfigure}{.5\textwidth} 
  \centering
  \includegraphics[width=.99\linewidth]{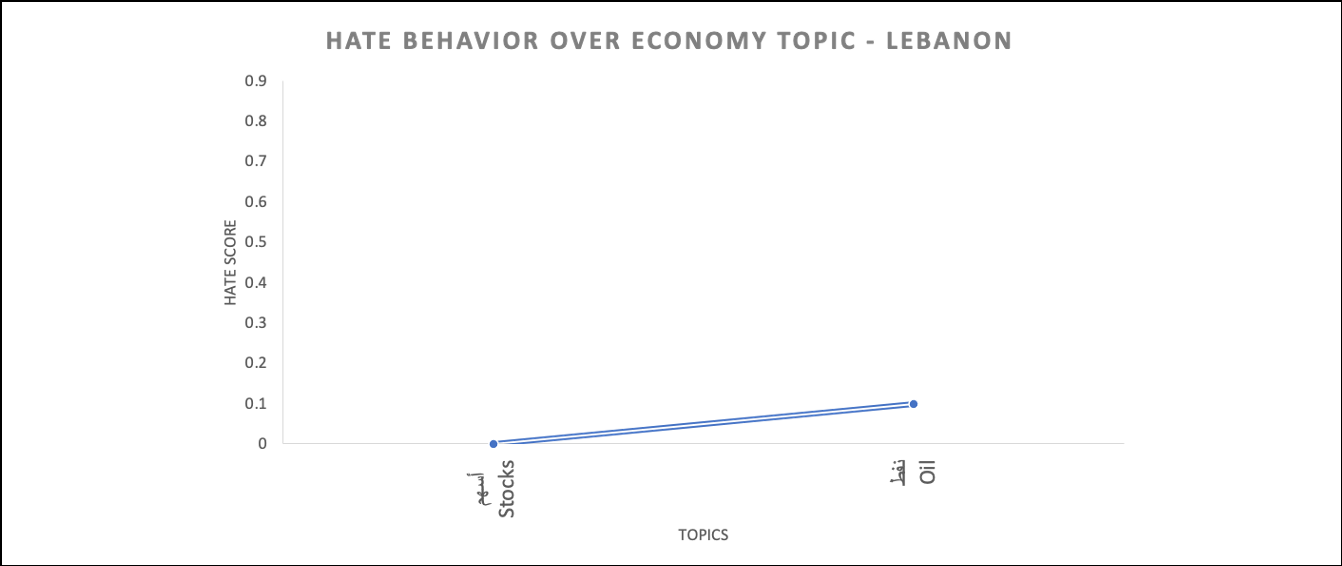}
  \caption{\scriptsize }
  \label{fig:leb-hate-economy-appendix}
\end{subfigure}
\begin{subfigure}{.5\textwidth} 
  \centering
  \includegraphics[width=.99\linewidth]{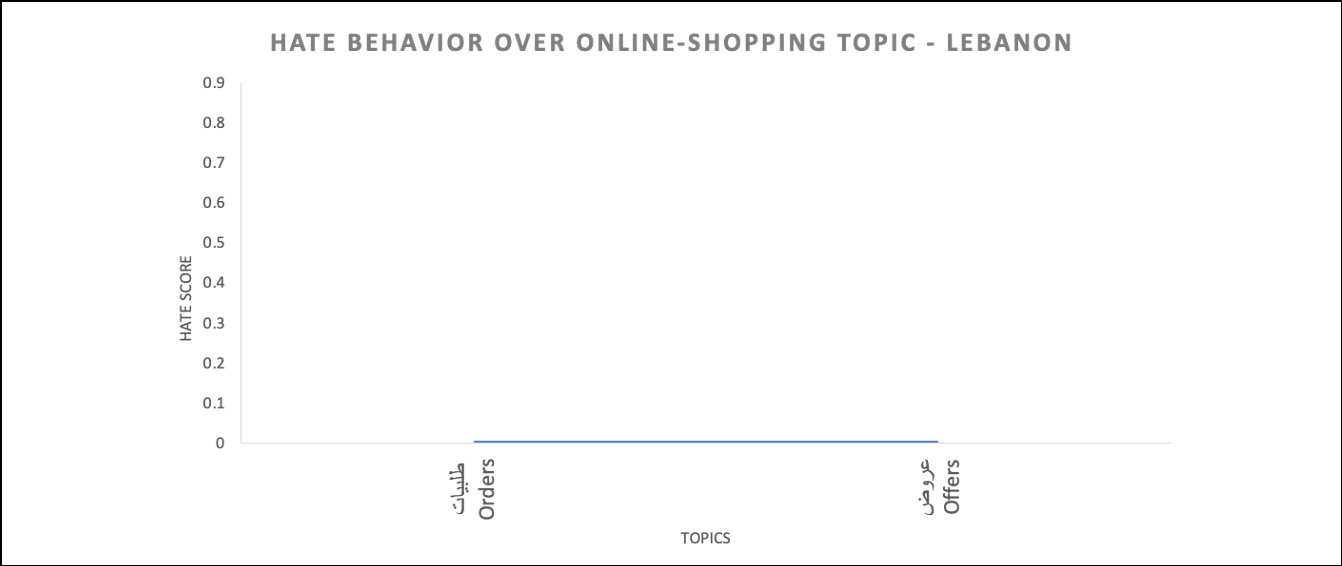}
  \caption{\scriptsize }
  \label{fig:leb-hate-onlineshopping-appendix}
\end{subfigure}%
\begin{subfigure}{.5\textwidth} 
  \centering
  \includegraphics[width=.99\linewidth]{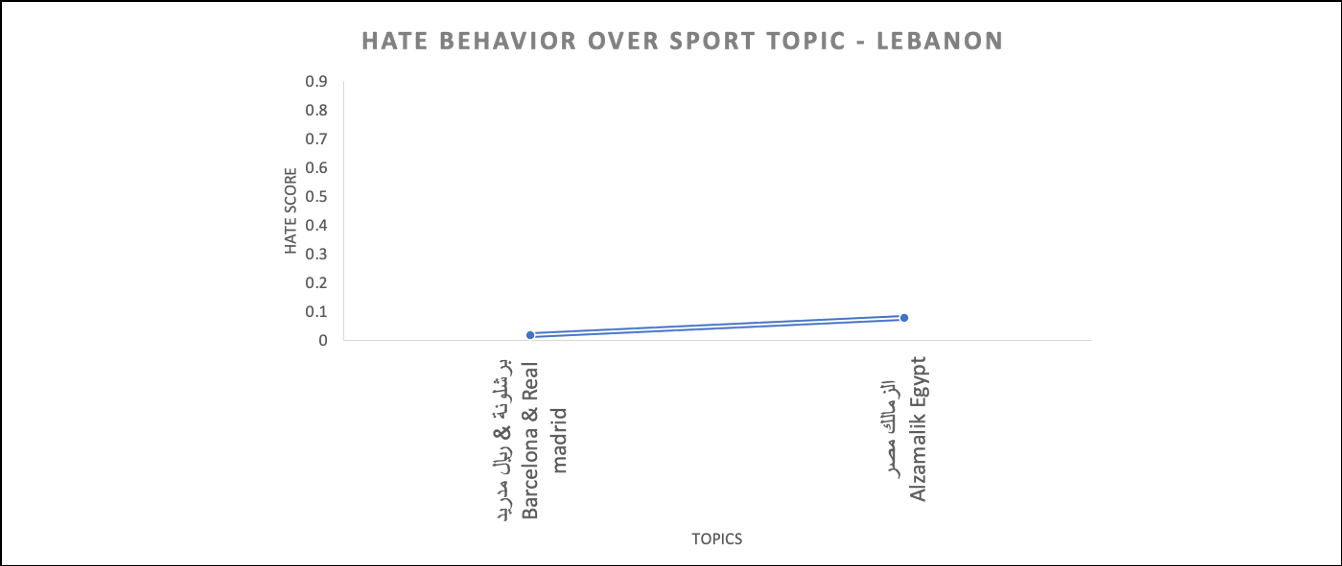}
  \caption{\scriptsize }
  \label{fig:leb-hate-sport-appendix}
\end{subfigure}
\caption{\small Subtopics of the topics inferred from Lebanon COVID-19 dataset for hate behavior analysis in Lebanon.}
\label{fig:leb-hate-detail-topics-appendix}
\end{figure}


\begin{figure}[h]
\centering
\begin{subfigure}{.5\textwidth} 
  \centering
  \includegraphics[width=.99\linewidth]{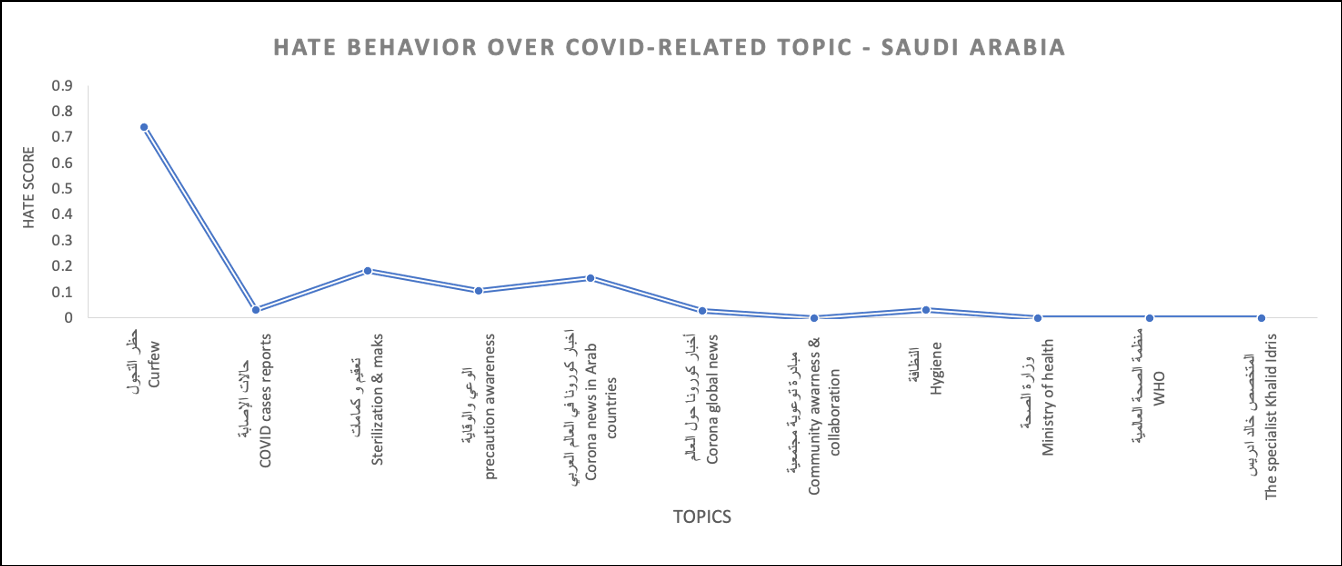}
  \caption{\scriptsize }
  \label{fig:saudi-hate-covid-appendix}
\end{subfigure}%
\begin{subfigure}{.5\textwidth} 
  \centering
  \includegraphics[width=.99\linewidth]{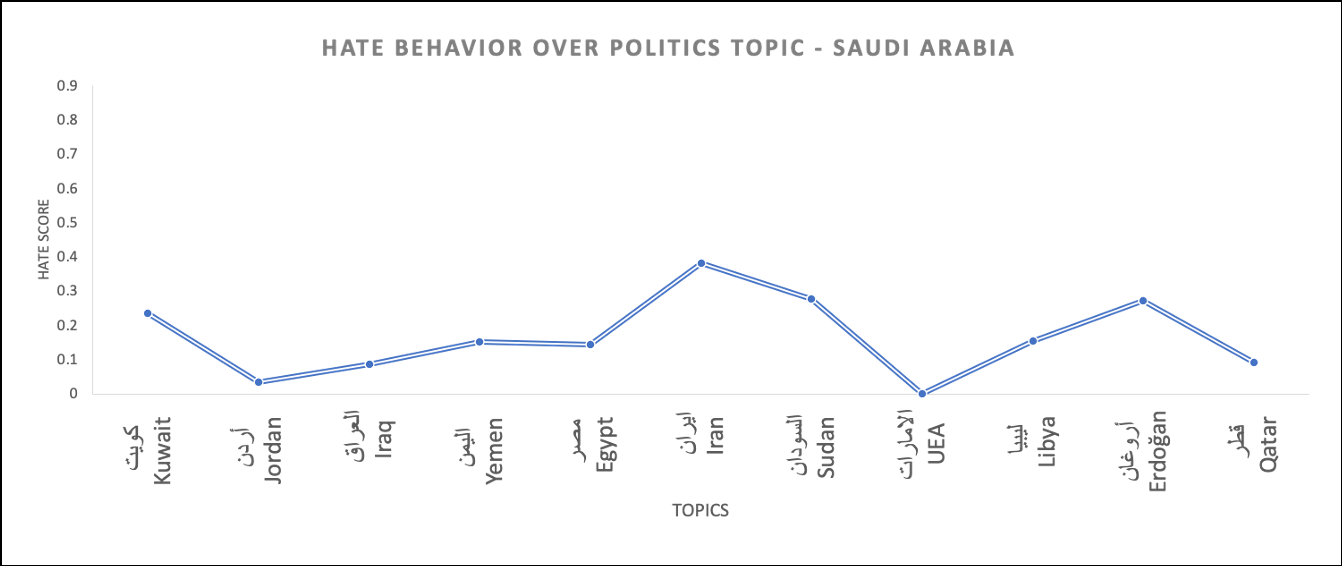}
  \caption{\scriptsize }
  \label{fig:saudi-hate-politics-appendix}
\end{subfigure}
\begin{subfigure}{.5\textwidth} 
  \centering
  \includegraphics[width=.99\linewidth]{images/saudi-sent-hate-topics/saudi-hate-local2.png}
  \caption{\scriptsize }
  \label{fig:saudi-hate-local-appendix}
\end{subfigure}%
\begin{subfigure}{.5\textwidth} 
  \centering
  \includegraphics[width=.99\linewidth]{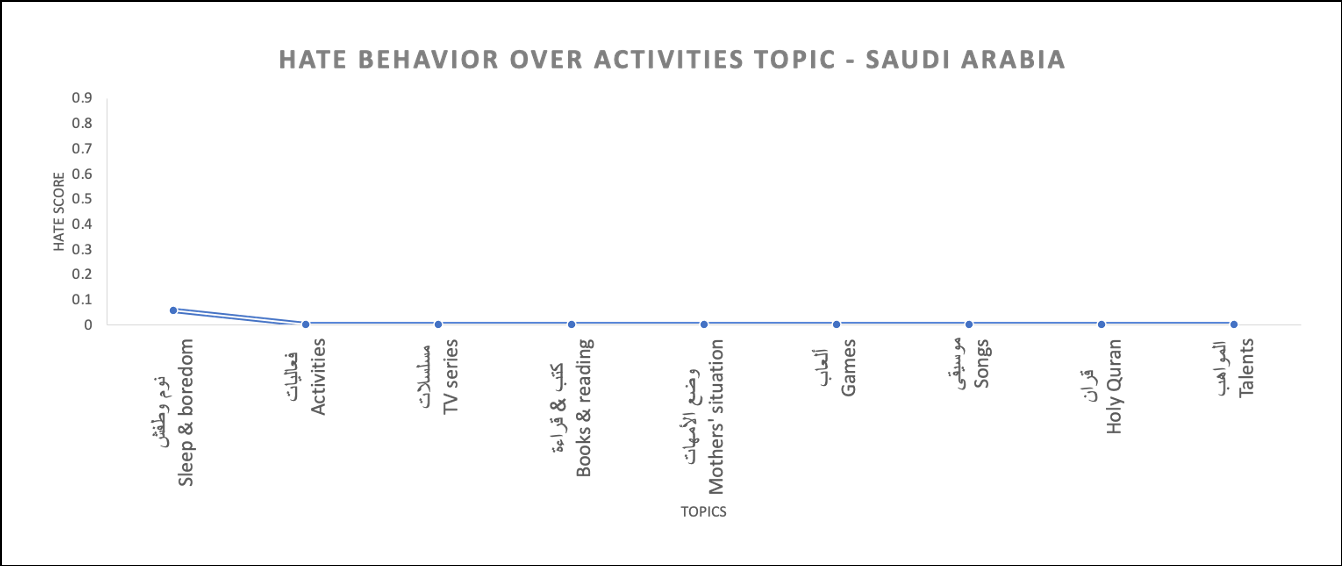}
  \caption{\scriptsize }
  \label{fig:saudi-hate-activities-appendix}
\end{subfigure}
\begin{subfigure}{.5\textwidth} 
  \centering
  \includegraphics[width=.99\linewidth]{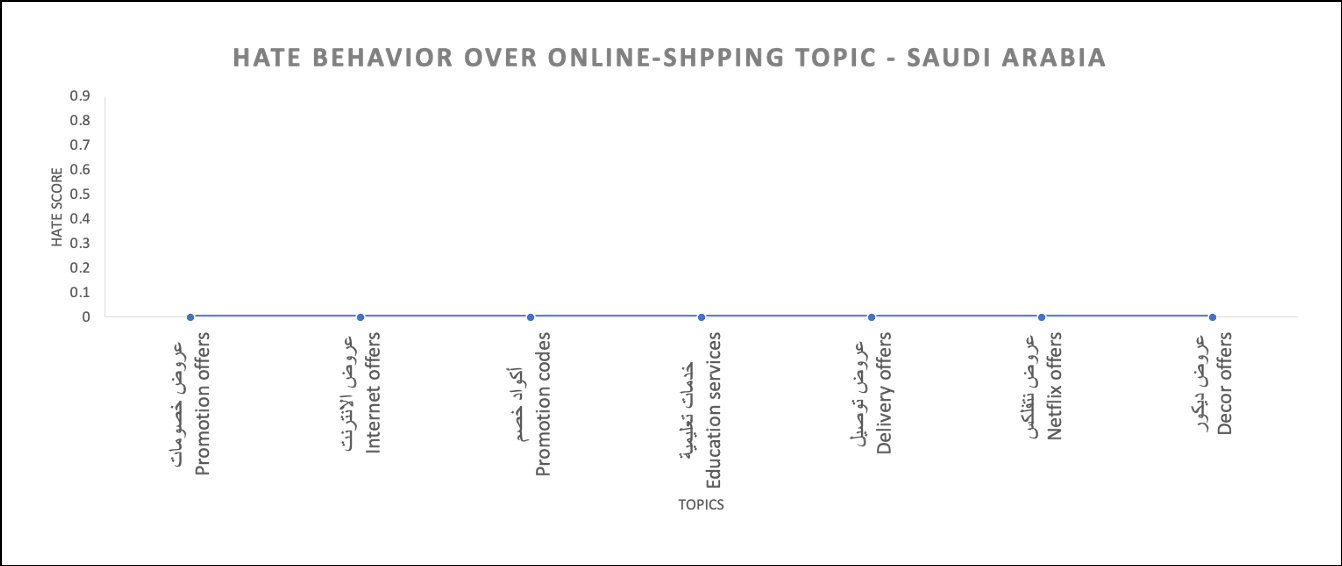}
  \caption{\scriptsize }
  \label{fig:saudi-hate-shopping-appendix}
\end{subfigure}%
\begin{subfigure}{.5\textwidth} 
  \centering
  \includegraphics[width=.99\linewidth]{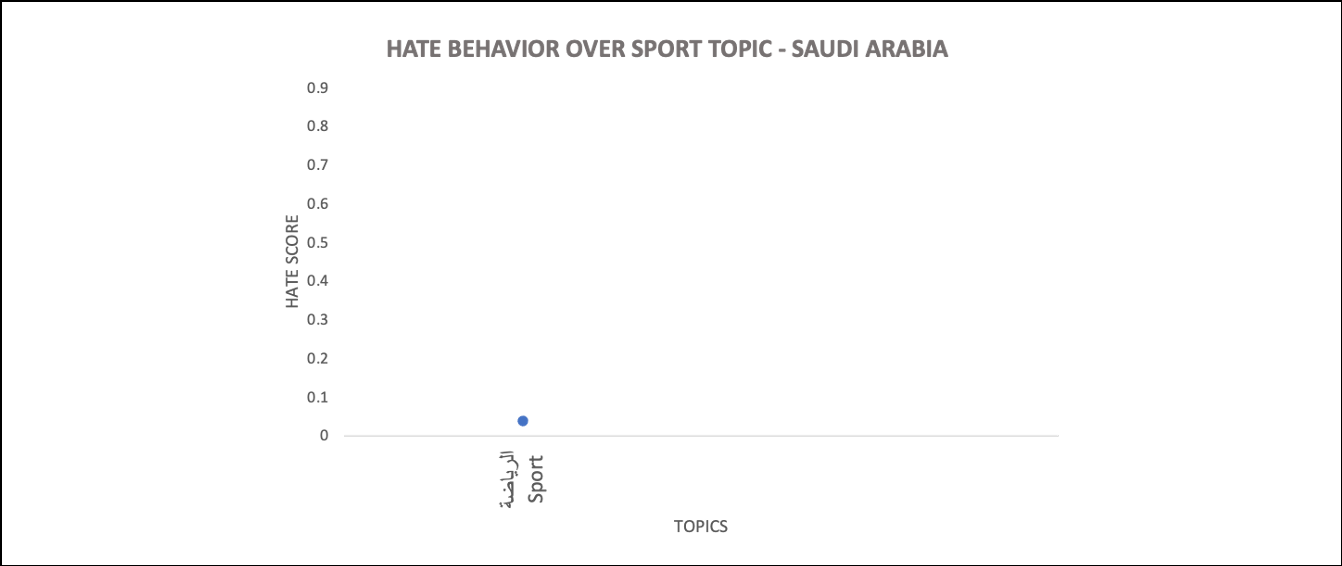}
  \caption{\scriptsize }
  \label{fig:saudi-hate-sport-appendix}
\end{subfigure}
\caption{\small Subtopics of the topics inferred from Saudi Arabia COVID-19 dataset for hate behavior analysis in Saudi Arabia.}
\label{fig:saudi-hate-detail-topics-appendix}
\end{figure}

\begin{landscape}
 \begin{figure}
\centering
\includegraphics[width=1.2\textwidth]{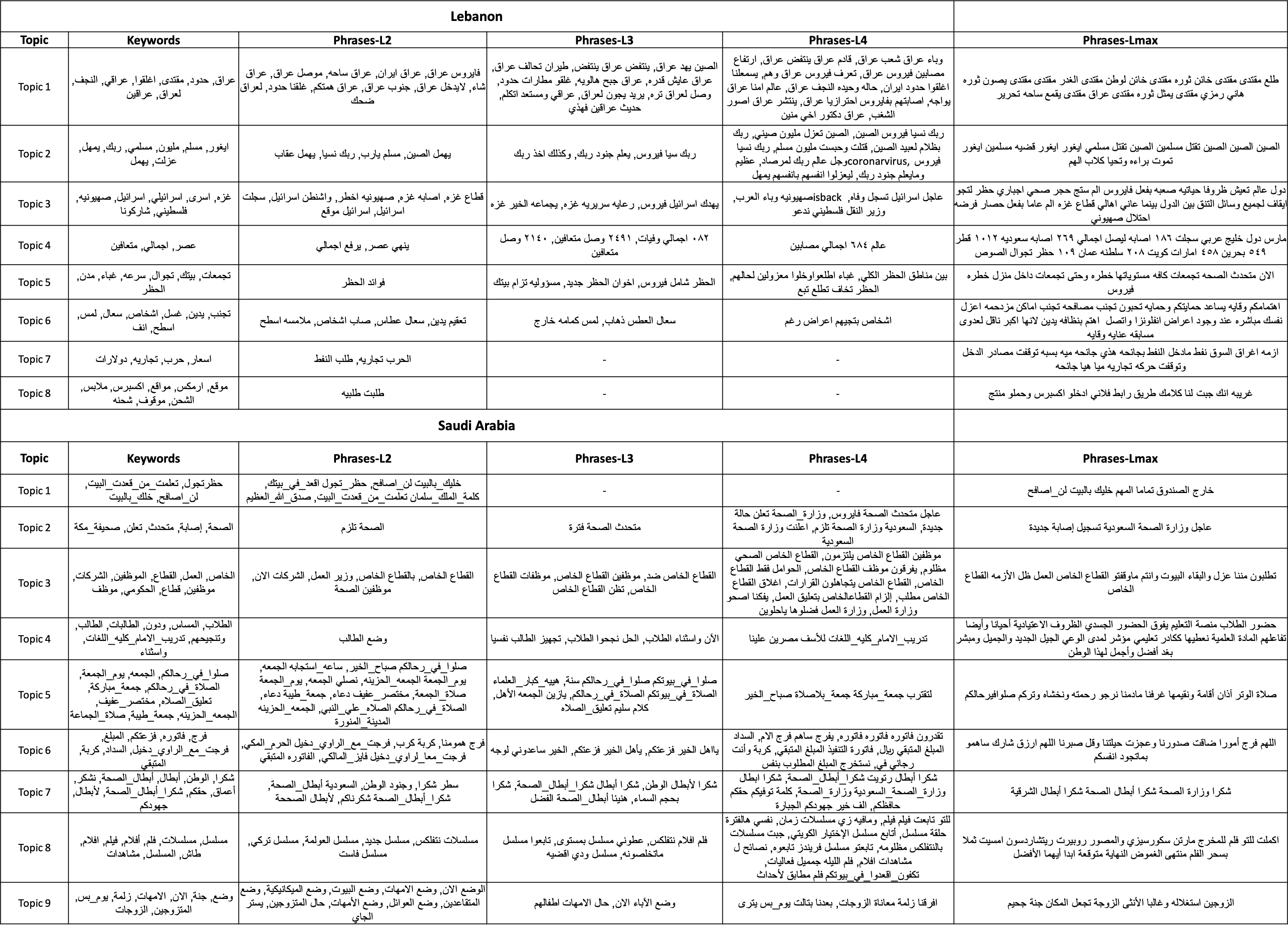}
\caption{Top keywords and phrases extracted using RAKE based on BERTopic top keywords. The keywords and phrases are ranked based on RAKE - Lebanon and Saudi Arabia.}
\label{fig:rake-phrases-leb-saudi-appendix}
\end{figure}
\end{landscape}


\bibliographystyle{unsrt}  
\bibliography{references}  

\end{document}